\documentclass[journal]{IEEEtran}

\usepackage{ifpdf}
\usepackage{multirow}
\usepackage{amssymb}

\usepackage{color}
\usepackage{xcolor}
\usepackage{subfigure}

\usepackage{amsmath}
\usepackage{algorithm}
\usepackage{algorithmic}
\usepackage{url}
\usepackage{cite}
\usepackage{array}

\ifCLASSINFOpdf
   \usepackage[pdftex]{graphicx}
   \DeclareGraphicsExtensions{.pdf,.jpeg,.png}
\else
\fi

\ifCLASSOPTIONcompsoc
 \usepackage[caption=false,font=normalsize,labelfont=sf,textfont=sf]{subfig}
\else
 \usepackage[caption=false,font=footnotesize]{subfig}
\fi

\newcommand{\black}[1]{\textcolor{black}{#1}}

\begin{document}
\title{Contrast-reconstruction Representation Learning for Self-supervised Skeleton-based Action Recognition}

\author{Peng Wang, Jun Wen, Chenyang Si, Yuntao Qian, \IEEEmembership{Senior Member, IEEE}, and Liang Wang, \IEEEmembership{Fellow, IEEE}

\thanks{This work was jointly supported by the National Key Research and Development Program of China under Grant 2018AAA0100500 and the National Natural Science Foundation of China under Grant 62071421. \emph{(Corresponding author: Yuntao Qian.)}}

\thanks{P. Wang and Y. Qian are with College of Computer Science, Zhejiang University, Hangzhou, Zhejiang 310007, China (e-mail: pengwang18@zju.edu.cn; ytqian@zju.edu.cn).}

\thanks{J. Wen is with Harvard Medical School. The work is done when he was at Zhejiang University. (E-mail: jungel2star@gmail.com).}

\thanks{C. Si and L. Wang are with Center for Research on Intelligent Perception and Computing (CRIPAC), National Laboratory of Pattern Recognition (NLPR), Institute of Automation, Chinese Academy of Sciences (CASIA) (e-mail: ChenYang.Si.Mail@gmail.com, wangliang@nlpr.ia.ac.cn).}}

\maketitle

\begin{abstract}
Skeleton-based action recognition is widely used in varied areas, \emph{e.g.}, surveillance and human-machine interaction. Existing models are mainly learned in a supervised manner, thus heavily depending on large-scale labeled data, which could be infeasible when labels are prohibitively expensive. In this paper, we propose a novel Contrast-Reconstruction Representation Learning network (CRRL) that simultaneously captures postures and motion dynamics for unsupervised skeleton-based action recognition. It consists of three parts: Sequence Reconstructor (SER), Contrastive Motion Learner (CML), and Information Fuser (INF). SER learns representation from skeleton coordinate sequence via reconstruction. However the learned representation tends to focus on trivial postural coordinates and be hesitant in motion learning. To enhance the learning of motions, CML performs contrastive learning between the representation learned from coordinate sequences and additional velocity sequences, respectively. Finally, in the INF module, we explore varied strategies to combine SER and CML, and propose to couple postures and motions via a knowledge-distillation based fusion strategy which transfers the motion learning from CML to SER. Experimental results on several benchmarks, \textit{i.e.}, NTU RGB+D 60/120, PKU-MMD, CMU, and NW-UCLA, demonstrate the promise of the our method by outperforming state-of-the-art approaches.
\end{abstract}



\section{Introduction}  \label{introduction}

\IEEEPARstart{A}{ction} recognition is a fundamental task in computer vision and is widely applied in robotics and human-computer interaction~\cite{weinland2011survey}. Though remarkable performances have been achieved in RGB video based action recognition~\cite{Ji20133DCN, Wang2012MiningAE, luo2017unsupervised, Wang2021TDNTD}, the high computational expense and vulnerability to background or viewpoint variations limit their applications in practice. Recently, due to the wide availability of depth sensors and significant developments of pose estimation methods, skeleton-based action recognition, which is much more computationally efficient and inherently robust against those variations, is receiving increasing research attention.

\begin{figure}
  \centering
  \subfigure[]{
  \includegraphics[width=0.229\linewidth]{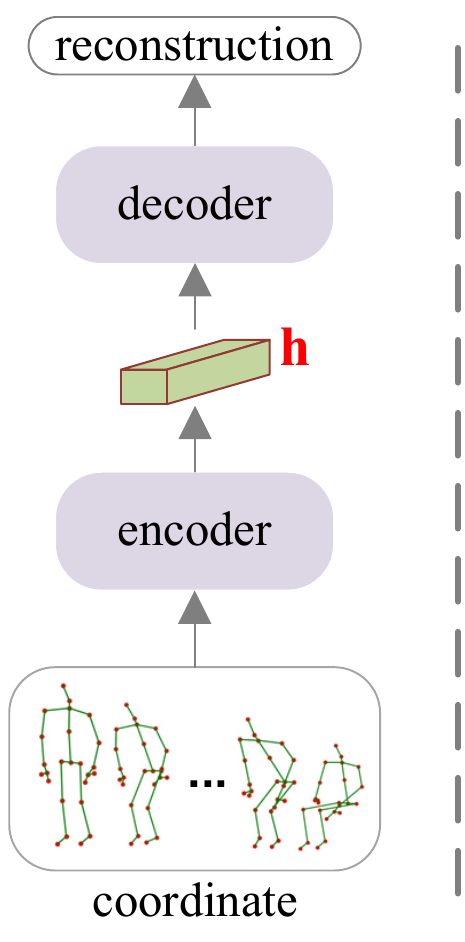}}
  \subfigure[]{
  \includegraphics[width=0.727\linewidth]{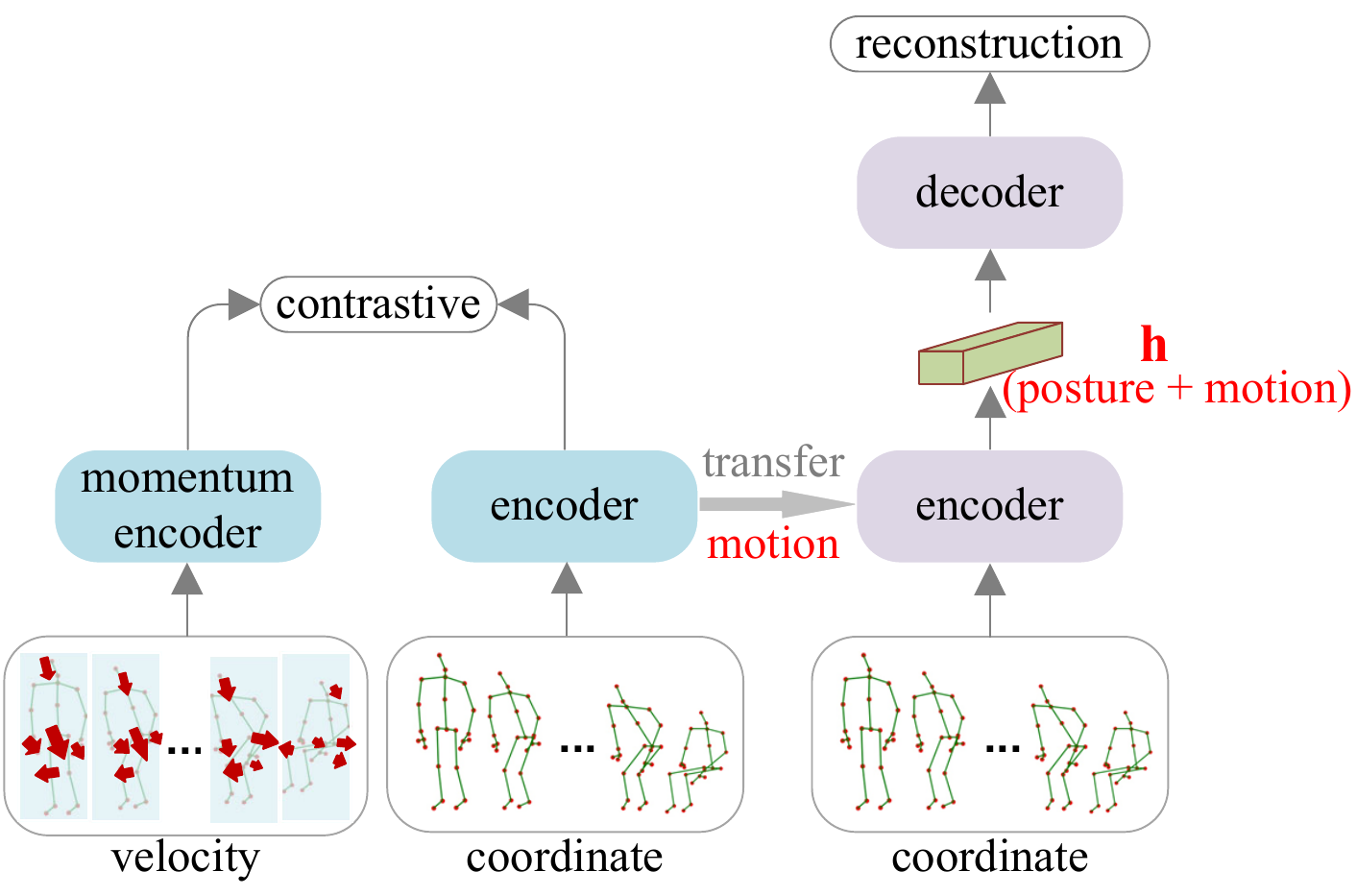}}
  \caption{Method comparison. (a) Existing methods are typically built on an auto-encoder architecture that receives skeleton coordinate sequences and learns representation via reconstruction. To best reconstruct the input, the learned representation tends to focus on trivial postural coordinate details, hesitant in learning motions. (b) By performing contrastive learning with the additional velocity information, we enhance the learning from skeletal coordinates of motions, which are then transferred to a typical auto-encoder to capture skeletal postures and motions simultaneously.}
  \label{fig_compare}
\end{figure}

The key to skeleton-based action recognition lies in the modeling of discriminative postures and temporal dynamics in skeleton sequences. To this end,  plenty of approaches have been proposed, \emph{i.e.}, two-stream networks~\cite{wang2017modeling, li2018co}, spatio-temporal attention based models~\cite{Song2018SpatioTemporalAL}, modified Long-Short Term Memory (LSTM) networks~\cite{liu2016spatio}, and integrated models~\cite{si2018skeleton}. Most of these methods are learned in a supervised manner, thus heavily relying on large-scale labeled training data, and are hardly feasible when labels are prohibitively expensive to be obtained.

\begin{figure*}[ht]
  \centering
  \includegraphics[width=0.9\linewidth]{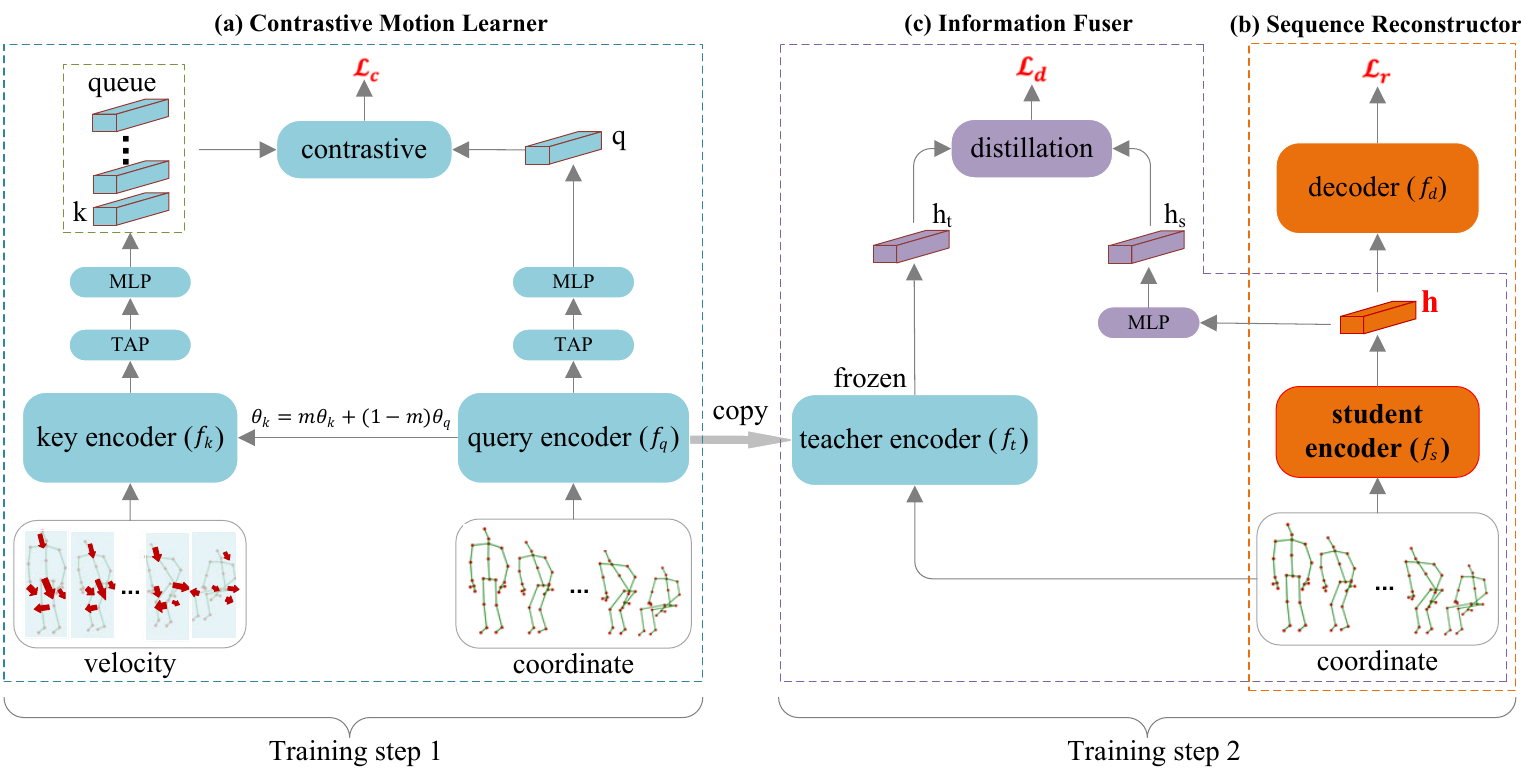}
  \caption{Architecture of the proposed Contrast-Reconstruction Representation Learning framework (CRRL). It consists of three parts: (a) the \emph{Contrastive Motion Learner} (CML) extracts motion dynamics by maximizing the high-level similarity between the velocity sequence and coordinate sequence; (b) the \emph{Sequence Reconstructor} (SER) learns postures via coordinate sequence reconstruction; (c) the \emph{Information Fuser} (INF) couples CML and SER via knowledge distillation --- the query encoder of CML plays teacher role and 'teaches' the student encoder of SER with motion dynamics via imposing the distillation loss. The learned representation of the student encoder thereby captures both the motion dynamics and postural information, which is then used for evaluation. The TAP in CML denotes temporal average pooling which is used for aggregating information across time~\cite{si2019attention}.}
\label{overview}
\end{figure*}

Recently, unsupervised skeleton-based action recognition is becoming more and more popular, with most of which are built on sequence prediction or reconstruction, as shown in Fig.~\ref{fig_compare}(a).~\cite{zheng2018unsupervised} proposed to learn action representation by inpainting the masked skeleton sequences.~\cite{su2020predict} presented a decoder-weakened auto-encoder that predicts skeleton sequences.~\cite{lin2020ms2l} presented a model containing three pretext tasks, \emph{i.e.}, skeleton prediction, jigsaw puzzle recognition, and contrastive learning. However, these methods are built upon the raw skeletal coordinates, which tend to guide the representation learning to focus on static postures, failing to capture the skeletal motion dynamics. As a result, some actions are hardly distinguishable via the representation, \emph{e.g.}, ``running'' and ``jogging''. To facilitate the modeling of motions, we propose to take advantage of explicit motion information, \emph{i.e.} velocity, into representation learning (see Fig.~\ref{fig_compare}(b)). This is motivated by the observation that, compared with raw skeletal coordinates, velocity inherently contains richer motion dynamics for classifying actions, which is also shown in~\cite{luo2017unsupervised, li2018co, cheng2021hierarchical, Wang2021TDNTD, Ng2018TemporalDN}. 

In this paper, we propose a novel unsupervised representation learning framework, called Contrast-Reconstruction Representation Learning (CRRL), for skeleton-based action recognition that incorporates additional skeletal velocity over raw coordinates to capture postures and motions simultaneously. As shown in Fig.~\ref{overview}, the framework is mainly composed of three parts: 1) \emph{Sequence Reconstructor} (SER); 2) \emph{Contrastive Motion Learner} (CML); and 3) \emph{Information Fuser} (INF). SER is of an encoder-decoder architecture, in which the encoder embeds skeletal coordinates sequences into fixed-dimensional representations for the decoder to reconstruct the input sequences. Since the auto-encoder is optimized to best reconstruct the original sequences, the learned representation tends to focus on detailed joint spatial coordinates, which is trivial for motion classification. To enhance the learning of motions from skeletal coordinates, we propose a CML module. It contains a query encoder and a key encoder that receives skeletal coordinate sequences and velocity sequences, respectively. The query encoder is guided to pay attention to motion learning from coordinate sequences via contrastive learning which maximizes the commonality, \emph{i.e.}, the motions, between the two encoders. 

With the SER and CML trained to learn postures and motions, respectively, we propose an INF module that enables the transfer of motion learning from CML to SER and couples them in a knowledge distillation manner~\cite{hinton2015distilling, yim2017gift}. Specifically, the query encoder of CML plays a ``teacher'' role guiding the encoder of SER to additionally pay attention to motion dynamics from raw skeletal coordinates. As a result, the learned representation by the encoder of SER captures motions and postures simultaneously.

We summarize the main contributions of our work as follows:

\begin{itemize}

  \item [$\bullet$] A novel unsupervised representation learning framework for skeleton-based action recognition is proposed, which simultaneously captures skeletal postures and motion dynamics by incorporating additional velocity information over the typically-adopted coordinate sequences.
  
  \item [$\bullet$] The learning ability of motion dynamics is promoted by performing contrastive learning between the representations that are learned from skeletal coordinate sequences and velocity sequences, respectively. By transferring motion learning via knowledge distillation, we fuse motion and posture in one representation that is learned from coordinate sequences.   

  \item [$\bullet$] Experiments on five benchmark datasets are conducted to validate the effectiveness of the proposed motion dynamic learning approach and information fusion strategy, which outperform state-of-the-art models by significant margins. 
\end{itemize}


\section{Related Works}  \label{related_works}

We first take a look into self-supervised visual representation learning methods. Then we review the works on supervised skeleton-based action recognition. At the end of this section, we present the recent advances in self-supervised skeleton-based action recognition.


\subsection{Self-supervised Visual Representation Learning}

Because labeled data is usually costly, self-supervised approaches for visual representation learning arouse increasing interest in the last several years. It constructs the optimization objective from the input data itself to train the model. According to the pretext tasks designed as the optimization objective, the self-supervised visual representation learning can be summarized into two main categories: generative and contrastive. The generative methods learn visual representation via solving pretext tasks that involve generating images or videos, measuring the loss in the output space. This type of methods includes image colorization~\cite{zhang2016colorful}, image inpainting~\cite{pathak2016context}, video reconstruction (or prediction)~\cite{srivastava2015unsupervised}, \textit{etc}. The contrastive methods are mostly based on the instance discrimination, \emph{i.e.} bringing the representations of different augmentations of the same instance closer while pushing the representations of different instances apart. They impose the contrastive loss, \emph{e.g.} InfoNCE~\cite{Oord2018RepresentationLW}, on representation space. In the last two years, many promising contrastive learning methods have  been proposed, such as SimCLR~\cite{chen2020simple}, MoCo~\cite{He2020MomentumCF}, BYOL~\cite{grill2020bootstrap}, and SimSiam~\cite{chen2021exploring}, \textit{etc}. Beyond the methods mentioned above, some other self-supervised learning approaches were based on context-related pretext tasks, \textit{e.g.} image jigsaw puzzle~\cite{noroozi2016unsupervised, kim2018learning}, video jigsaw puzzle~\cite{ahsan2019video}, temporal coherency~\cite{mobahi2009deep}, temporal order verification~\cite{misra2016shuffle}, the arrow of time recognition~\cite{wei2018learning}, \textit{etc}. For more comprehensive reviews on self-supervised learning, the readers can refer to~\cite{jing2020self, liu2021self}.


\subsection{Supervised Skeleton-based Action Recognition}

Before the prevalence of deep learning, traditional skeleton-based action recognition models were mainly built upon hand-crafted action descriptors. For example,~\cite{ohn2013joint} proposed two descriptors, with one for joint angle similarities and the other for modified HOG. ~\cite{evangelidis2014skeletal} proposed a local skeleton descriptor to encode relative positions of joint quadruples.~\cite{vemulapalli2014human} proposed to map actions from the Lie group to its Lie algebra and performed classification using a combination of dynamic time warping. 

With the development of deep learning, deep neural network based methods become prevalent and significant advancements have been achieved~\cite{du2015hierarchical,Song2018SpatioTemporalAL, li2017skeleton, si2019attention,yan2018spatial, li2019actional, zhu2016co, shahroudy2016ntu, zhang2019view, li2018co, liu2020trajectorycnn, yan2018spatial, si2018skeleton,shi2019two, Si2019AnAE, chen2021multi, liu2016spatio}. Recurrent Neural Network (RNN) based methods were first investigated. \cite{liu2016spatio} proposed to simultaneously model skeletal joints' spatial and temporal dependencies via a spatio-temporal LSTM network. \cite{Song2018SpatioTemporalAL} proposed a spatial and temporal attention model that is capable of selectively focusing on discriminative joints of skeletons and paying different levels of attention to different frames. \cite{zhu2016co} designed a co-occurrence regularized deep LSTM network. \cite{shahroudy2016ntu} introduced a part-aware LSTM to model interactions of different human parts. \cite{zhang2019view} designed a view adaptive LSTM to enable adapting to the most suitable observation viewpoint. Then, by treating joints coordinates sequence as a matrix, convolutional network based methods have been proposed~\cite{li2018co,liu2020trajectorycnn}. Recently, by treating the skeletons as graph nodes, graph neural network based methods have been proposed~\cite{yan2018spatial,si2018skeleton,shi2019two, Si2019AnAE, chen2021multi}. Despite their success, these methods are heavily dependent on labeled data, which could be infeasible in practice due to the high expense of labels.


\subsection{Self-supervised Skeleton-based Action Recognition}

Self-supervised skeleton-based action recognition methods aim to learn discriminative features from the skeleton sequence without human annotations. They can be roughly divided into two categories: self-reconstruction based learning and contrastive learning.

For self-reconstruction based learning,~\cite{zheng2018unsupervised} proposed a seq2seq model to inpaint the masked part of the sequence. To avoid the inpainted sequence being blurry and visually unrealistic, a discriminator is adopted to distinguish between real and fake sequences. With a RNN based encoder-decoder,~\cite{su2020predict} developed two strategies to weaken the decoder and strengthen the representation learned by its encoder.~\cite{xu2020prototypical} proposed to create reverse sequential predictions with action prototypes that encode semantic similarity implicitly shared among sequences. Recently,~\cite{yang2021skeleton} treated a skeletal sequence as a skeleton cloud and colorized each point in the cloud according to its temporal and spatial orders in the original skeleton sequence, and then this skeleton cloud was repainted to extract semantic information for action recognition.

For contrastive learning,~\cite{rao2021augmented} proposed to perform contrastive learning among different augmentations of unlabeled skeleton data.~\cite{thoker2021skeleton} presented a framework that learns invariance from different skeleton representations, \emph{i.e.} graph, sequence, and image representation. By processing the skeleton data in multiple views, \emph{i.e.} joint, motion and bone,~\cite{Li2021Human} introduced a cross-view contrastive learning framework that can leverage multi-view complementary supervision signal.~\cite{su2021self} proposed to leverage motion consistency to construct positive pairs for skeletal contrastive learning.

A most relevant work~\cite{lin2020ms2l} integrated three pretext tasks for unsupervised representation learning, including motion prediction, jigsaw puzzle recognition, and contrastive learning. The main difference between this work and ours is that we explicitly fuse motion and postural representation that are learned separately. Specifically, 1) beyond the representation learning directly on skeletal coordinate space in \cite{lin2020ms2l}, we explicitly guide motion learning by performing contrastive learning between skeletal coordinates and skeletal velocity. Since velocity sequences are computed as the differences between two consecutive coordinate frames, and thus the absolute spatial coordinate information is lost, the model is forced to focus on commonalities in temporal dynamics to maximize the similarity. 2) we alleviate the potential incompatibility between motion and postural learning and fuse them in a fine-grained manner by distilling them onto one joint representation.


\section{The Proposed Method}

In this section, we present the proposed unsupervised representation learning framework for skeleton-based action recognition. As illustrated in Fig.~\ref{overview}, it consists of three parts: SER, CML and INF. SER is an RNN based auto-encoder, mainly learning postures from skeletal coordinate sequences. CML promotes the capturing of motion dynamics by performing contrastive learning between the representations learned from coordinate sequences and velocity sequences, respectively. INF couples the postures learned from SER with the motion dynamics learned from CML via a knowledge distilling strategy.


\subsection{Sequence Reconstructor}

The \emph{Sequence Reconstructor} (SER) consists of an encoder and a decoder, both with Gated Recurrent Units (GRUs) layers. The encoder reads in the skeleton sequences and maps them to representations, which are then fed into the decoder to reconstruct the input sequences.

Specifically, let $\mathcal{X}=\{x_i\}_{i=1}^N$ denotes the training set of $N$ skeleton sequences, and each sequence $x=\{x^t\}_{t=0}^T \in \mathbb{R}^{T \times J \times 3}$ has $T$ frames, with each frame containing $J$ joints. The encoder runs through a skeleton sequence and encodes it into a sequence of hidden states denoted as $h=\{h^t\}_{t=0}^T$. We repeat the last state $h^T$ for $T$ times to form a sequence and then feed it into the decoder as ``hints'' to help the decoder reconstruct the input sequences, as done in~\cite{zheng2018unsupervised}, which is shown to strengthen the importance of the learned representation with improved performance. We use bidirectional GRUs as the encoder to capture richer dynamics and thus feed the forward-pass and backward-pass representation of the encoder simultaneously to the decoder which is uni-directional GRUs.

\begin{figure}[ht]
    \centering
    \includegraphics[width=0.99\linewidth]{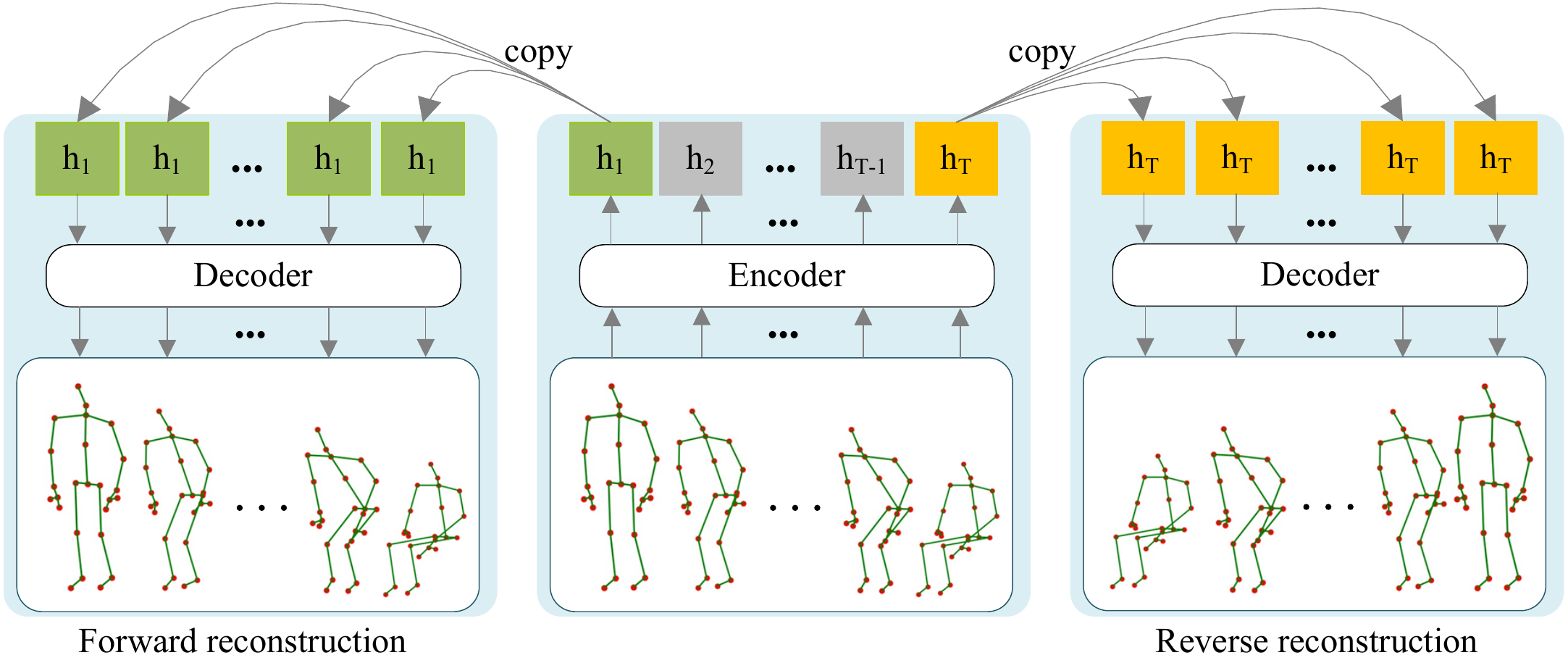}
    \caption{The proposed SER reconstructs the input skeleton sequence both forwardly and reversely. Specifically, it repeats the first hidden state $T$ times as a sequence fed into decoder to reconstruct the input skeleton sequence forwardly, and repeats the last hidden state $T$ times as a sequence fed into decoder to reconstruct the input skeleton sequence reversely.}
  \label{recon-both-direction}
  \end{figure}

With the learned representation, the decoder is trained to reconstruct the input sequence both forwardly and reversely, as shown in Fig.~\ref{recon-both-direction}. It is motivated by the observation in ~\cite{srivastava2015unsupervised,zheng2018unsupervised} that with uni-directional reconstruction only, the encoder and decoder tend to couple together and learn trivial representation by just remembering the input sequences. We observe that requiring the decoder to replay the skeleton sequences in both directions guides the encoder to learned representation that focuses on skeleton dynamics. Specifically, we use the mean squared error (MSE) as the reconstruction loss $\mathcal{L}_{r}$:  
\begin{equation}
\mathcal{L}_{r} = ||x - \hat x||_2^2 + ||\overleftarrow{x} - \hat{\overleftarrow{x}}||_2^2,
\label{eq:reconstrction_loss}
\end{equation}
\noindent where $\overleftarrow{x}$ denotes the reversed input sequence; $\hat x$ and $\hat{\overleftarrow{x}}$ denote the forwardly and reversely reconstructed sequences, respectively.


\subsection{Contrastive Motion Learner}
In this section, we aim to promote motion learning from skeletal coordinate sequences by performing contrastive learning between coordinate sequences and velocity sequences. Specifically, the velocity sequence is defined as the differences between two consecutive coordinate frames,
\begin{equation}
v^t =
\begin{cases}
x^{t+1} - x^t,  & \text{if $0 \leq t < T$} ; \\
v^{t-1}, & \text{if $t =T$,}
\end{cases}
\label{compute-velocity}
\end{equation}

\noindent where $T$ denotes the sequence length. Then a velocity sequence is represented by $v=\{v^t\}_{t=0}^T$. \black{Since velocity inherently contains skeletal dynamics, we assume that if the encoder extracts motions from raw skeletal coordinates, the learned representation would be consistently similar to the one learned from the corresponding velocity sequence and distinct from others. To achieve this, we perform contrastive representation learning between coordinate sequences and velocity sequences. Specifically, following~\cite{He2020MomentumCF}, we denote a coordinate representation as the query and its corresponding velocity representation as the positive key and the other velocity representation as negative keys. The encoder is trained to maximize the similarity between the query and the positive key and the dissimilarity between the query and the negative keys.}

As shown in Fig.~\ref{overview}(a), the proposed \emph{Contrastive Motion Learner} (CML) consists of two GRU encoders, a query encoder and a key encoder. The query encoder receives coordinate sequences while the key encoder receives velocity sequences. To aggregate temporally global action information, we apply Temporal Average Pooling (TAP) to sequential hidden states~\cite{Si2019AnAE, rao2021augmented}. A nonlinear projection~\cite{chen2020mocov2, chen2020simple} is performed right after the TAP, which prevents information loss caused by contrastive loss. Following~\cite{chen2020mocov2}, we perform nonlinear projection using a Multi-layer Perceptron (MLP) head which consists of two fully connected layers. Thus, the query representation is:
\begin{equation}
\label{compute-q}
q = MLP_{q}(\frac{1}{T}\sum_{t=0}^T \it{f}_{\it{q}}(\it{x}^t)),
\end{equation}
where $MLP_{q}(\cdot)$ denotes the MLP head on the top of query encoder, and $f_q(\cdot)$ is the transformation of the query encoder. Likewise, the key representation is:
\begin{equation}
\label{compute-k}
k = MLP_{k}(\frac{1}{T}\sum_{t=0}^T \it{f}_{\it{k}}(\it{v}^t)),
\end{equation}
where $MLP_{k}(\cdot)$ denotes the MLP head on the top of key encoder, and $f_k(\cdot)$ denotes the key encoder.

At each training step, the current mini-batch of skeleton sequences and their corresponding velocity sequences are fed into the query encoder and key encoder, respectively. The extracted query representations and key representations form positive pairs, and thus these key representations are denoted as $k_+$. Following \cite{He2020MomentumCF}, to perform efficient contrastive learning, we maintain a queue of key representations which is updated with new keys at each training step. All the existing key representations in the queue are viewed as negative keys $k_-$ for the training of the next mini-batch. To mitigate the inconsistency of key representations, the key encoder is updated in a momentum manner~\cite{He2020MomentumCF}:
\begin{equation}
\label{momentum-update}
\theta_k = m\theta_k + (1-m)\theta_q,
\end{equation}
where $\theta_k$, $\theta_q$ denote the parameters of key encoder and query encoder, respectively. $m \in [0, 1)$ is a momentum coefficient. Only the parameters $\theta_q$ are updated by back-propagation. The momentum update makes $\theta_k$ evolve more smoothly than $\theta_q$.

CML promotes the learning of motion dynamics residing in the coordinate sequences. For each coordinate query representation $q$, the positive velocity key $k_+$ is trained to be similar to $q$ while the negative keys $k_-$ to be dissimilar to $q$. With similarity measured by dot product, the encoder is trained via an InfoNCE objective~\cite{Oord2018RepresentationLW}, which is defined as:
\begin{equation}
\label{contr_loss}
\mathcal{L}_{c}=-\log \frac{\exp \left(q \cdot k_{+} / \tau\right)}{\sum_{i=0}^{K} \exp \left(q \cdot k_{i} / \tau\right)},
\end{equation}
where $\tau$ is a temperature hyper-parameter that controls the concentration level of the similarity distribution~\cite{Wu2018UnsupervisedFL}. CML maximizes the similarity between the query representation and the one positive key over $K$ negative keys.


\subsection{Information Fuser}

In this section, we couple the postures learned by SER with the motion dynamics learned from CML via knowledge distillation. Specifically, we treat the query encoder of CML as a teacher that transfers its motion learning capability to the encoder of SER via knowledge distillation such that the SER encoder learns from skeletal coordinate sequence a representation that captures motion dynamics and postures simultaneously. We observe that such a fine-grained knowledge fusion approach achieves better performance than direct feature concatenation or pre-training as shown in Section IV-D1.

As shown in Fig. \ref{overview} (c), given a skeleton sequence as input, the teacher encoder gets $h_t$ while the student encoder (\textit{i.e.}, the SER encoder) gets $h$. We find that if regularization is directly performed on the student encoder's $h$, its reconstruction role could be corrupted due to its space scale being different from the teacher encoder. Thus, we first project $h$ into $h_s$ using two fully-connected layers. Then, the $L2$ similarity between $h_s$ and $h_t$ is utilized as the feature distillation loss for information transfer, formulated as:
\begin{equation}
\label{eq:distillation_loss}
\mathcal{L}_{d} = ||h_t - h_s||_2^2.
\end{equation}


\subsection{Network Training}

Algorithm 1 provides training procedure of the proposed network. We first train CML to minimize the objective in Eq. (\ref{contr_loss}) to guide the learning of motion dynamics. Then we train SER and INF with a joint loss: 
\begin{equation}
\mathcal{L}_{joint} = \mathcal{L}_{r} + \lambda_d \mathcal{L}_{d},
\end{equation}

\noindent where $\lambda_d$ is a hyper-parameter that balances the reconstruction objective (Eq. (\ref{eq:reconstrction_loss})) and information fusion objective (Eq. (\ref{eq:distillation_loss})). To be noted, during the joint training, the parameters of the teacher encoder from CML are fixed, and only the parameters of SER and the nonlinear projection are updated. Through the joint training, we aim to transfer the learned motions by CML to SER and capture postures and motions simultaneously for action recognition.

\begin{figure}[!t]
	\label{algor}
	\renewcommand{\algorithmicrequire}{\textbf{Input:}}
	\renewcommand{\algorithmicensure}{\textbf{Output:}}
	\begin{algorithm}[H]
		\caption{Training of CRRL}
		\begin{algorithmic}[1]
			\REQUIRE Dataset $\mathcal{D}$, momentum $m$, temperature $\tau$, queue size $K$, epoch $e$ for the first training step, epoch $e'$ for the second training step, $\lambda_d$. 
            
            \STATE randomly initialize the query encoder $f_q$, the student encoder $f_{s}$, the SER decoder $f_{d}$, the $MLP$ and the $queue$;
            \STATE copy the parameters of $f_q$ to the key encoder $f_k$;
			
			\hspace*{\fill} \\
			\STATE \textcolor{gray}{\# Step 1: training CML}
			\FOR{epoch $\in$ $e$}{
			\FOR{$x \in$ mini-batch}{
			\STATE compute $v$ by Eq. (\ref{compute-velocity});
			\STATE compute $q$ by Eq. (\ref{compute-q});
			\STATE compute $k$ by Eq. (\ref{compute-k});
			\STATE stop gradients to $k$;
			\STATE $l_{pos} = q \cdot k$; \  \textcolor{gray}{\# positive logit: $1 \times 1$}
			\STATE $l_{neg} = q \cdot queue$; \  \textcolor{gray}{\# negtive logits: $1 \times K$}
			\STATE $l_{all} = concatenate(l_{pos}, l_{neg})$   \textcolor{gray}{\# all logits: $1 \times (K+1)$}
			\STATE $loss = CrossEntropy(l_{all}, 0)$; \textcolor{gray}{\# the positive is the 0\-th}
			}
			\ENDFOR
			\STATE update $f_q$ by back-propagation;
			\STATE update $f_k$ by Eq. (\ref{momentum-update});
			\STATE enqueue $k$ to $queue$;
			\STATE dequeue the oldest mini-batch of keys;
			}
			\ENDFOR \quad  \textcolor{gray}{\# optimum query encoder $f_q$ is achieved}

			\hspace*{\fill} \\
			\STATE \textcolor{gray}{\# Step 2: training SER and INF}
			\FOR{epoch $\in$ $e'$}{
			\FOR{$x \in$ mini-batch}{
			\STATE $\boldsymbol{h} = f_{s}(x)$;  \  \textcolor{gray}{\# h: hidden embedding}
			\STATE $\hat x = f_{d}(h_1)$;   
			\STATE $\overleftarrow{\hat{x}} = f_{d}(h_T)$;
			\STATE $\overleftarrow{x} = reverse(x)$;
			\STATE $\mathcal{L}_{r} = ||x - \hat x||_2^2 + ||\overleftarrow{x} - \overleftarrow{\hat{x}}||_2^2$; \  \textcolor{gray}{\# reconstruction loss}
			
			\STATE $h_s = MLP(TAP(h))$;
			\STATE $f_t = f_q$ with require\_grad=false; \ \textcolor{gray}{\# no gradient to $f_t$}
			\STATE $h_t = TAP(f_{t}(x))$;
			\STATE $\mathcal{L}_{d} = ||h_t - h_s||_2^2$; \ \textcolor{gray}{\# distillation loss}
			\STATE $\mathcal{L}_{joint} = \mathcal{L}_{r} + \lambda_d \mathcal{L}_{d}$;
			}
			\ENDFOR
			\STATE update $f_s, f_d, MLP$ by back-propagation;
			}
			\ENDFOR

			\ENSURE optimum student encoder $f_s$.

		\end{algorithmic}
	\end{algorithm}
\end{figure}


\section{Experiments}

\subsection{Datasets}

We perform the experiments on five published datasets: NTU RGB+D 60, NTU RGB+D 120, PKU-MMD, CMU, and NW-UCLA. We briefly describe them as follows.

\textbf{NTU RGB+D 60 Dataset~\cite{shahroudy2016ntu}}. This is a large-scale action recognition dataset containing 60 different action classes, including daily, mutual, and health-related actions. It consists of 56,880 video samples performed by 40 distinct subjects. These samples are recorded by Microsoft Kinect v2 cameras and have four modalities: depth maps, 3D skeleton sequence, RGB frames, and IR sequences. The skeleton of each body contains 3D coordinates of 25 joints. This dataset was captured from 3 different views: front view, left side 45 degrees view, and right side 45 degrees view. There are two standard evaluation protocols for this benchmark: cross-subject and cross-view. We test our method on both of them. 

\textbf{NTU RGB+D 120 Dataset~\cite{liu2019ntu}} extends NTU RGB+D 60 dataset by adding another 60 classes and another 57,600 video samples. This dataset is collected from 106 performers. It consists of 32 setups with different camera locations. Two evaluation protocols are provided: cross-subject and cross-setup. In the cross-subject protocol, 63,026 samples recorded by 53 subjects are used for training, while the left 50,919 samples are for testing. In cross-setup evaluation, the samples with even collection setup IDs are for training, and those with odd setup IDs are for test.

\begin{table}[t]
  \caption{Performance comparison on the NTU RGB+D 60 dataset with linear evaluation protocol (S: supervised, U: unsupervised).}
  \label{ntu60_linear}
  \centering
  \setlength{\tabcolsep}{4.1mm}{
  \begin{tabular}{p{0.05cm}|l|cc}
    \hline
    \multicolumn{2}{c|}{Methods}
    & C-View & C-Sub \\
    \hline
    \multirow{5}{*}{S}
    & HOPC~\cite{rahmani2014hopc} & 52.8 & 50.1 \\
    & HBRNN~\cite{du2015hierarchical} & 64.0 & 59.1 \\
    & Part-Aware-LSTM~\cite{shahroudy2016ntu} & 70.3 & 62.9 \\
    & ST-LSTM~\cite{liu2016spatio} & 77.7 & 69.2 \\
    & VA-RNN-Aug~\cite{zhang2019view} & 87.6 & 79.4 \\
    
    \hline
    \multirow{6}{*}{U}
    & LongT GAN~\cite{zheng2018unsupervised} & 56.4 & 52.1 \\
    & MS$^2$L~\cite{lin2020ms2l} & - & 52.6 \\
    & PCRP~\cite{xu2020prototypical} & 63.5 & 53.9 \\
    & CAE+~\cite{rao2021augmented} & 64.8 & 58.5 \\
    & 3s-CrosSCLR(LSTM)~\cite{Li2021Human} & 69.2 & 62.8 \\
    & CRRL (\textbf{Ours})  & \textbf{73.8} & \textbf{67.6} \\
  
  \hline
\end{tabular}}
\end{table}

\begin{table}[t]
  \caption{Performance comparison on the NTU RGB+D 120 dataset with linear evaluation protocol (S: supervised, U: unsupervised).}
  \label{ntu120}
  \setlength{\tabcolsep}{4.1mm}{
  \centering
  \begin{tabular}{p{0.05cm}|l|cc}
    \hline
    \multicolumn{2}{c|}{Methods}
    & C-Setup & C-Sub \\
    \hline
    \multirow{3}{*}{S}
    & Soft RNN~\cite{hu2018early} & 44.9 & 36.3 \\
    & ST-LSTM~\cite{liu2016spatio} & 57.9 & 55.7 \\
    & RotClips+MTCNN~\cite{ke2018learning} & 61.8 & 62.2 \\
    \hline
    \multirow{4}{*}{U}
    & LongT GAN~\cite{zheng2018unsupervised} & 39.7 & 35.6 \\
    & PCRP~\cite{xu2020prototypical} & 45.1 & 41.7 \\
    & CAE+~\cite{rao2021augmented} & 49.2 & 48.6 \\
    & 3s-CrosSCLR(LSTM)~\cite{Li2021Human} & 53.2 & 53.9 \\
    & CRRL (\textbf{Ours})  & \textbf{57.0} & \textbf{56.2} \\

  \hline
\end{tabular}}
\end{table}

\begin{table}[t]
  \caption{Performance comparison on the PKU-MMD dataset with linear evaluation protocol (S: supervised, U: unsupervised).}
  \label{pkummd-table}
  \centering
  \setlength{\tabcolsep}{5mm}{
  \begin{tabular}{p{0.05cm}|l|cc}
    \hline
    \multicolumn{2}{c|}{Methods}
    & Part I & Part II \\
    \hline
    \multirow{3}{*}{S}
    & SA-LSTM~\cite{Song2018SpatioTemporalAL} & 86.3 & - \\
    & TA-LSTM~\cite{Song2018SpatioTemporalAL} & 86.6 & - \\
    & STA-LSTM~\cite{Song2018SpatioTemporalAL} & 86.9 & - \\
    \hline
    \multirow{4}{*}{U}
    & P\&C FW-AEC~\cite{su2020predict} & 59.9 & 25.5 \\
    & MS$^2$L~\cite{lin2020ms2l} & 64.9 & 27.6 \\
    & LongT GAN~\cite{zheng2018unsupervised} & 67.7 & 26.0 \\
    & CRRL (\textbf{Ours})  & \textbf{82.1} & \textbf{41.8} \\

  \hline
\end{tabular}}
\end{table}

\begin{table}[ht]
  \caption{Performance comparison on the CMU dataset with linear evaluation protocol (S: supervised, U: unsupervised).}
  \label{cmu}
  \centering
  \setlength{\tabcolsep}{4mm}{
  \begin{tabular}{p{0.05cm}|l|c}
    \hline
    \multicolumn{2}{c|}{Methods} & Accuracy(\%)\\
    \hline
    \multirow{2}{*}{S} 
    & HBRNN~\cite{du2015hierarchical} & 75.0 \\ 
    & Co-deep LSTM~\cite{zhu2016co} & 81.0 \\
    \hline
    \multirow{3}{*}{U}
    & Composite LSTM~\cite{srivastava2015unsupervised} & 61.4  \\
    & LongT GAN~\cite{zheng2018unsupervised} & 66.2 \\
    & CRRL (\textbf{Ours})  & \textbf{78.7} \\

  \hline
\end{tabular}}
\end{table}

\begin{table}[ht]
  \caption{Performance comparison on the NW-UCLA dataset with linear evaluation protocol (S: supervised, U: unsupervised).}
  \label{nw-ucla}
  \centering
  \setlength{\tabcolsep}{4mm}{
  \begin{tabular}{p{0.05cm}|l|c}
    \hline
    \multicolumn{2}{c|}{Methods} & Accuracy(\%)\\
    \hline
    \multirow{3}{*}{S} 
    & HOPC~\cite{rahmani2014hopc} & 74.2 \\
    & HBRNN-L~\cite{du2015hierarchical} & 78.5 \\
    & VA-RNN-Aug~\cite{zhang2019view} & 90.7 \\
    \hline
    \multirow{3}{*}{U}
    & LongT GAN~\cite{zheng2018unsupervised} & 74.3 \\
    & MS$^2$L~\cite{lin2020ms2l} & 76.8 \\
    & CRRL (\textbf{Ours})  & \textbf{83.8} \\

  \hline
\end{tabular}}
\end{table}

\begin{table}[ht]
  \caption{Performance comparison on the NW-UCLA dataset and NTU RGB+D 60 dataset with KNN classifier (U: unsupervised).}
  \label{knn}
  \centering
  \setlength{\tabcolsep}{3mm}{
  \begin{tabular}{c|l|c|cc}
    \hline
    \multicolumn{2}{c|}{\multirow{2}{*}{Methods} }
    & \multirow{2}{*}{NW-UCLA} 
    & \multicolumn{2}{c}{NTU 60} \\
    \cline{4-5}
    \multicolumn{2}{c|}{} &  & C-View & C-Sub \\
    \hline
    \multirow{4}{*}{U}
    & LongT GAN~\cite{zheng2018unsupervised} & 74.3 & 48.1 & 39.1 \\
    & P\&C FS-AEC~\cite{su2020predict} & 83.8 & \textbf{76.3} & 50.6 \\
    & P\&C FW-AEC~\cite{su2020predict} & 84.9 & 76.1 & 50.7 \\
    & CRRL (\textbf{Ours})  & \textbf{86.4} & 75.2 & \textbf{60.7} \\

  \hline
\end{tabular}}
\end{table}

\begin{table}[ht]
  \caption{Performance comparison of transfer learning.}
  \label{transfer-table}
  \centering
  \setlength{\tabcolsep}{4mm}{
  \begin{tabular}{c|cc}
    \hline
    \multirow{2}{*}{Methods}                  & \multicolumn{2}{c}{Transfer to PKU-MMD Part II} \\
    \cline{2-3}
                                              & PKU-MMD Part I & NTU RGB+D 60 \\
    \hline
    LongT GAN~\cite{zheng2018unsupervised}    & 43.6   & 44.8 \\
    MS$^2$L~\cite{lin2020ms2l}                & 44.1   & 45.8 \\
    CRRL (\textbf{Ours})                      & \textbf{47.0} & \textbf{48.5} \\

  \hline
\end{tabular}}
\end{table}

\begin{table*}[ht]
  \caption{Comparison of different fusion strategies.}
  \label{fusion-table}
  \centering
  \setlength{\tabcolsep}{4.1mm}{
  \begin{tabular}{p{2.5cm}|cc|cc|c|c}
    \hline

    \multirow{2}{*}{Fusion strategies}
    & \multicolumn{2}{c|}{NTU 60}
    & \multicolumn{2}{c|}{NTU 120}
    & \multirow{2}{*}{CMU} 
    & \multirow{2}{*}{NW-UCLA} \\
    \cline{2-5}
    & C-View    & C-Sub    & C-Setup    & C-Sub  \\
    \hline
    Rand-Enc          & 51.0   & 45.5   & 31.5   & 29.5   & 57.2   & 50.5  \\
    CML                & 68.1   & 62.8   & 51.6   & 50.9   & 76.7   & 72.8 \\
    SER               & 72.2   & 65.8   & 55.2   & 54.1   & 76.9   & 81.4 \\
    \hline
    CML-SER-joint      & 72.5   & 66.6   & 56.3   & 55.4   & 77.3   & 82.2 \\
    CML-pretrain-SER   & 72.0   & 66.5   & 55.4   & 54.4   & 76.0   & 80.8 \\
    SER-pretrain-CML   & 68.5   & 58.7   & 50.7   & 48.1   & 73.3   & 79.7 \\
    SER-distill-CML    & 71.9   & 64.6   & 54.6   & 53.6   & 77.4   & 80.6 \\
    CRRL  & \textbf{73.8}  & \textbf{67.6}  & \textbf{57.0} & \textbf{56.2} & \textbf{78.7} & \textbf{83.8}\\

  \hline
\end{tabular}}
\end{table*}

\textbf{PKU Multi-Modality Dataset (PKU-MMD)~\cite{Liu2020ABD}} is a large-scale benchmark for human action analytics with almost 20,000 action samples and 5.4 million frames. It consists of two parts, \textit{i.e.} Part I and Part II. Part I contains 1,076 untrimmed video sequences with 51 action classes performed by 66 subjects, while Part II contains 1,009 untrimmed video sequences with 41 action classes performed by 13 subjects. Compared to Part I, Part II is more challenging for action recognition because of large view variation and heavy occlusion. Following~\cite{lin2020ms2l}, we conduct experiments under the cross-subject protocol on Part I and Part II, respectively.

\textbf{CMU Dataset~\cite{cmu2003}}. Unlike NTU datasets recorded by Microsoft Kinect v2, the CMU dataset is captured by a Vicon motion capture system, and the actors wear markers and a black garment. Therefore, the noise in this dataset is much smaller. Nevertheless, action recognition is still very challenging because of the large sequence length variations and intra-class diversities in this dataset. It contains 2235 skeleton sequences categorized into 45 action classes~\cite{zhu2016co}. Each frame consists of 3D coordinates of 31 joints. We follow the experimental protocol proposed in~\cite{zheng2018unsupervised} and conduct four-fold cross validation in experiments.

\textbf{NW-UCLA Dataset~\cite{wang2014cross}} contains 1494 videos covering 10 actions. Each action is performed by 10 subjects and is captured by Kinect cameras from three viewpoints. Following~\cite{wang2014cross, liu2017enhanced}, we use the samples from the first and second viewpoint as training data while samples from the third viewpoint as test data.


\subsection{Implementation Details}

We first describe the sequence pre-processing steps. For NTU RGB+D 60/120 and PKU-MMD datasets, the skeleton sequences are downsampled to no more than 60 and 80 frames, respectively. Then we implement the coordinate translation proposed by~\cite{shahroudy2016ntu}: fixing the $X$ axis parallel to the 3D vector from “right shoulder” to “left shoulder”, and the $Y$ axis towards the 3D vector from “spine base” to “spine”. The $Z$ axis is fixed as the new $X \times Y$. For the CMU and NW-UCLA datasets, the skeleton sequences are downsampled to have at most 100 and 50 frames, respectively. Then, for all datasets, each action sample is separately normalized to $[-1,1]$ via min-max normalization, and each dimension in an action sample is normalized individually, \textit{i.e.,} the minima and maxima are computed per dimension across all frames in each sequence. For these action instances that involve two persons, we normalize each person's coordinates individually.

We implement the model in PyTorch and optimize it using SGD with weight decay of $1e^{-4}$ and SGD momentum of 0.9. All the encoders are two-layer bidirectional GRUs, while the decoder is two-layer unidirectional GRUs. All GRUs have 256 hidden units. Inside the encoders and the decoder, we use dropout between layers as proposed in~\cite{Zaremba2014RecurrentNN}, and the drop rate is 0.2. We train the networks on one NVIDIA Tesla V100 with the batch size of 32 for all experiments.

In the first training step, we train CML for 100 epochs, with learning rate of 0.01, 0.01, 0.005, 0.001, and 0.005 for CMU, NW-UCLA, NTU RGB+D 60, NTU RGB+D 120, and PKU-MMD datasets, respectively. The queue size K is 64 for all datasets (as will be discussed in Section~\ref{Analysis_CML_section}). We set temperature $\tau=0.1$. The momentum coefficient $m$ is 0.999. 

In the second training step, We train SER and INF for 60 epochs. The learning rate is initialized with 0.01, 0.04, 0.05, and 0.05 for CMU, NW-UCLA, NTU RGB+D 60/120, and PKU-MMD datasets, respectively, and reduced by multiplying it by 0.5 at 20 and 50 epochs.

When testing, the feature before the MLP head is used as the action representation~\cite{chen2020simple}. Specifically, when testing CML, the encoding from the query encoder is used as the action representation. When testing CRRL, the encoding from the student encoder is used as the action representation.


\subsection{Results and Comparison}   \label{comparison_section}

To verify the effectiveness of the representation learned by the proposed CRRL framework, we compare it with state-of-the-art models on action classification task, including the following recently-proposed unsupervised methods: LongT GAN~\cite{zheng2018unsupervised}, P\&C FW-AEC ~\cite{su2020predict}, MS$^2$L~\cite{lin2020ms2l}, PCRP~\cite{xu2020prototypical}, CAE+~\cite{rao2021augmented}, and 3s-CrosSCLR~\cite{Li2021Human}. These methods have been introduced in Section~\ref{related_works}. We first follow the widely used linear evaluation protocol~\cite{He2020MomentumCF, chen2020simple} and K-nearest neighbors (KNN) protocol~\cite{su2020predict} to implement the comparison experiments. Then, we conduct the across-dataset experiments to evaluate whether the representation learned by CRRL is transferrable. Unless specified, we report the action classification accuracy in percentage.


\subsubsection{Comparison under Linear Evaluation Protocol}

Following~\cite{He2020MomentumCF, chen2020simple}, we add a linear classifier on top of the student encoder of the CRRL to classify actions. The encoder's parameters are fixed when training the linear classifier. 

Table~\ref{ntu60_linear} compares our proposed method with previous supervised and unsupervised methods applied to NTU RGB+D 60 dataset. Our CRRL network outperforms the first three supervised approaches and achieves better performance (\textgreater 4\%) than previous unsupervised methods on both cross-view and cross-subject protocols. 

Table~\ref{ntu120} shows the comparison on the large-scale NTU RGB+D 120 dataset. Our CRRL model outperforms the supervised model Soft RNN~\cite{hu2018early} by a large margin (\textgreater 20\%) and even achieves comparable performance to ST-LSTM~\cite{liu2016spatio}. CRRL achieves better performance than the state-of-the-art by nontrivial margins on both cross-setup and cross-subject protocols.

The results on the PKU-MMD dataset are shown in Table~\ref{pkummd-table}. We improve the performance upon the compared unsupervised methods from 67.7\% to 82.1\% on Part I and from 27.6\% to 41.8\% on Part II. Our proposed unsupervised method achieves slightly lower performance than the supervised methods proposed in~\cite{Song2018SpatioTemporalAL}.

For the CMU dataset, although it is challenging for action recognition due to the large sequence length variations and intra-class diversities, the performance of our CRRL method is impressive and only slightly lower than the performance of the supervised method HBRNN~\cite{du2015hierarchical}, as shown in Table~\ref{cmu}. Again, our approach significantly improves over the state-of-the-art unsupervised learning methods.

For the NW-UCLA dataset, our CRRL model defeats the HOPC~\cite{rahmani2014hopc} and HBRNN-L~\cite{du2015hierarchical} supervised models and surpasses the state-of-the-art unsupervised learning methods by large margins.


\subsubsection{Comparison via KNN Classifier}

Following~\cite{su2020predict}, we compare our model to the state-of-the-art methods by using a KNN classifier on top of the frozen encoder. Specifically, we use a KNN classifier (K=1) to assign a class to each test sequence according to the cosine similarity between its features and the training sequences' features. Table~\ref{knn} shows the comparison on the NW-UCLA dataset and NTU RGB+D 60 dataset. Our approach improves over the state-of-the-art on the NW-UCLA dataset. For NTU RGB+D 60 dataset, our model achieves better performance than~\cite{su2020predict} in cross-subject protocol and comparable performance in cross-view protocol.


\subsubsection{Transfer Learning}

In this part, we evaluate the transfer learning performance of our model via across-dataset experiments. We follow the protocol used in~\cite{lin2020ms2l} which regards the NTU RGB+D 60 and PKU-MMD Part I as source datasets and PKU-MMD Part II as the target dataset. The proposed CRRL model is first pre-trained on sources datasets respectively in an unsupervised way. Then the student encoder along with a linear classifier are jointly fine-tuned on the target dataset with annotations. The transfer learning performance is evaluated on PKU-MMD Part II via action classification accuracy. Since the transfer learning performance of LongT GAN~\cite{zheng2018unsupervised} and MS$^2$L~\cite{lin2020ms2l} had been reported in the literature, we compare our model with them. As shown in Table~\ref{transfer-table}, our method surpasses these methods by about 3\%, demonstrating that the representations learned by our CRRL model are relatively more transferrable than those of existing models.


\subsection{Model Analysis}

In this section, we conduct experiments to give more analysis of the proposed approach. Firstly, we investigate different strategies for the combination of CML and SER in subsection~\ref{combination_section}. Then, subsection~\ref{vs} compares the proposed velocity-coordinate contrastive learning to the commonly used coordinate-coordinate contrastive learning. Subsection~\ref{visual} presents the visualizations of the action classifying results of CML, SER, and CRRL. Subsection~\ref{computational_costs} evaluates the computational cost of CML, SER, and CRRL. In subsection~\ref{Analysis_SER_section} and~\ref{Analysis_CML_section}, we conduct the module-level ablation study of CML and SER. Finally, in subsection~\ref{VG}, we propose another approach to leverage the velocity sequence for the learning of motion dynamics, and compare it with CML and CRRL. All the experiments in this section are performed with the linear evaluation protocol.


\subsubsection{Different Fusion Strategies}   \label{combination_section}

\begin{figure*}
  \centering
  \subfigure[Training loss of CML, SER, and CML-SER-joint]{
  \includegraphics[width=0.38\linewidth]{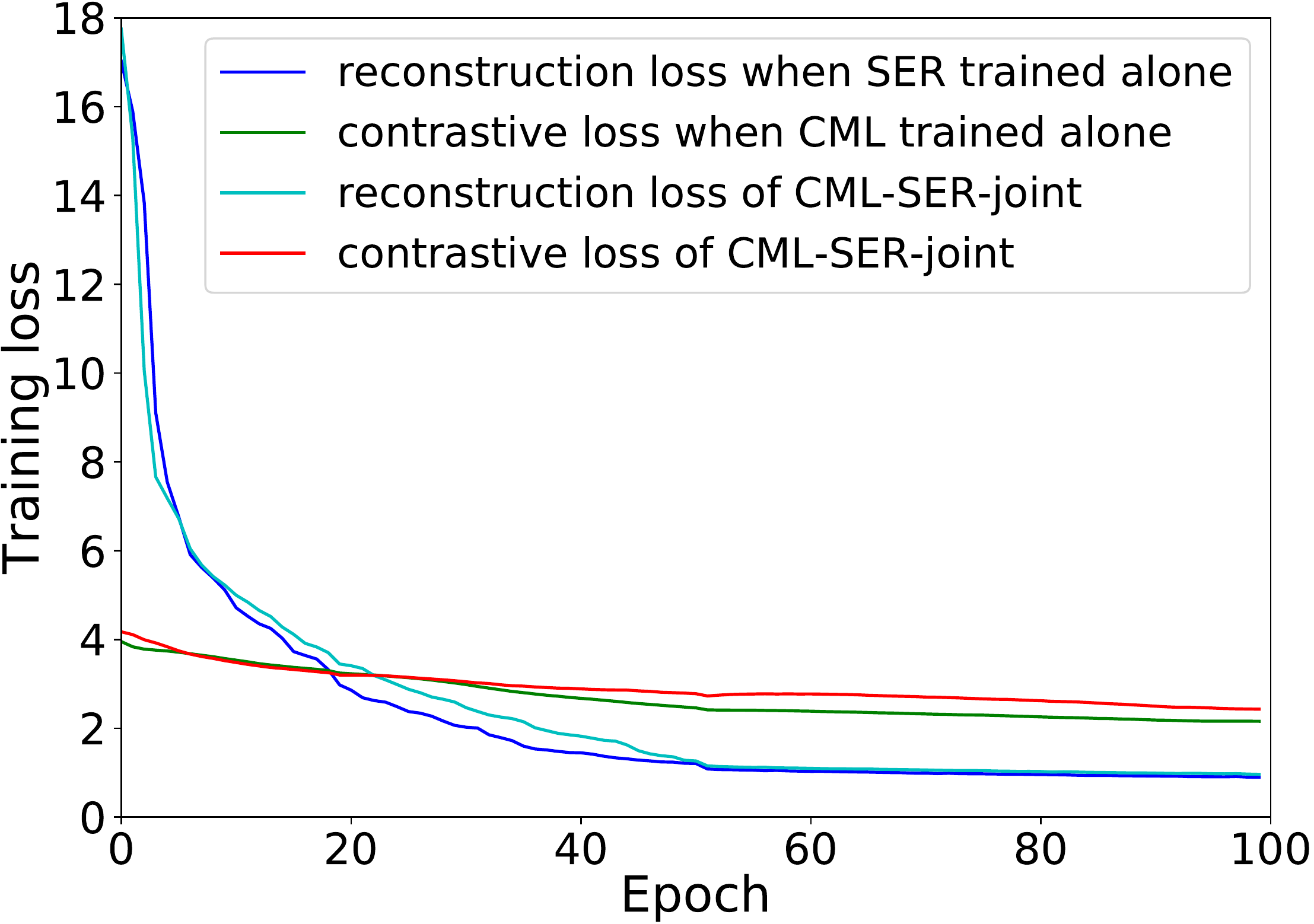}}
  \subfigure[Training loss of CRRL]{
  \includegraphics[width=0.38\linewidth]{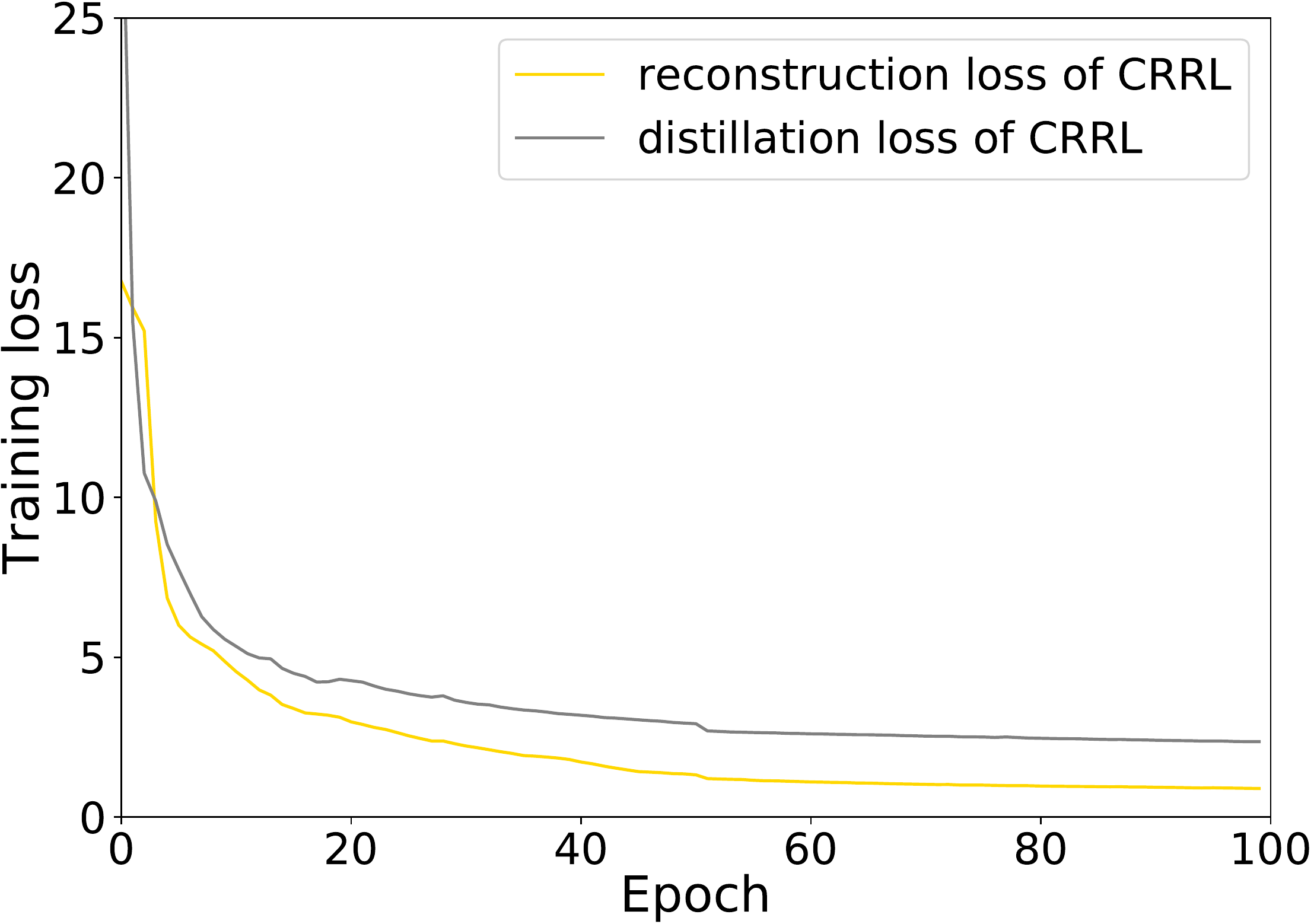}}
  \caption{The training loss of (a) CML, SER, CML-SER-joint, and (b) CRRL on NTU RGB+D 60 dataset with cross-subject protocol.}
  \label{loss}
\end{figure*}

\begin{table}
  \caption{Comparison of different augmentations for contrastive learning.}
  \label{contrastive-comparison}
  \centering
  \setlength{\tabcolsep}{2.6mm}{
  \begin{tabular}{l|cc|cc}
    \hline
    \multirow{2}{*}{Augmentations}
    & \multicolumn{2}{c|}{NTU 60}
    & \multicolumn{2}{c}{NTU 120}\\
    \cline{2-5}
    & C-View    & C-Sub    & C-Setup   & C-Sub   \\
    \hline
    Rand-Enc                 & 51.0     & 45.5   & 31.5  & 29.5 \\
    \hline
    Coordinate gaussian noise           & 54.8     & 49.6   & 34.9   & 36.9  \\
    Coordinate shear                    & 56.2     & 51.4   & 42.3   & 43.1  \\
    Coordinate rotation                 & 56.9     & 51.7   & 42.5   & 44.0  \\
    Velocity (\textbf{the proposed})  & \textbf{68.1}   & \textbf{62.8}  & \textbf{51.6}   & \textbf{50.9}\\
  \hline
\end{tabular}}
\end{table}

\begin{figure}
  \centering
  \subfigure[CML-SER-joint strategy]{
  \includegraphics[width=0.435\linewidth]{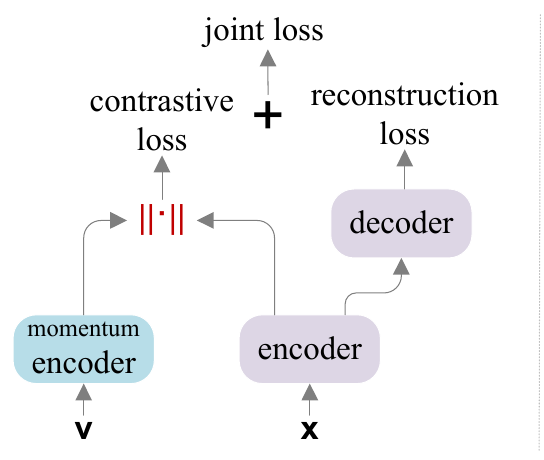}}
  \subfigure[SER-distill-CML strategy]{
  \includegraphics[width=0.525\linewidth]{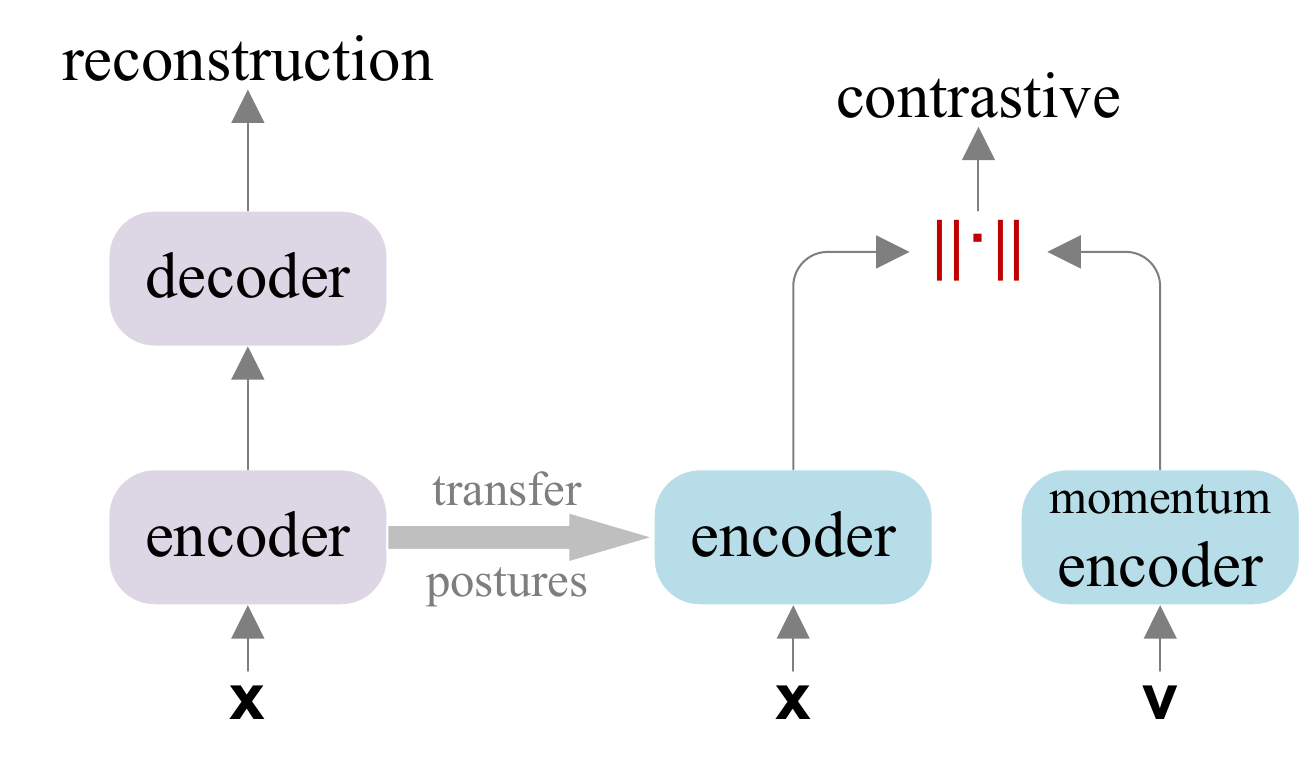}}
  \caption{Illustration of the information fusion strategies. (a) CML-SER-joint strategy: CML and SER are jointly optimized. (b) SER-distill-CML strategy: the learned knowledge of SER is distilled to CML. Namely, this approach is the reverse of CRRL.}
  \label{fusion_strategy}
\end{figure}

Proper fusion of the motion dynamics learned from CML with the postures learned from SER is critical for action recognition. In this section, we empirically study the fusion methods in depth. In addition to the proposed CRRL method, we also investigate another four approaches as follows:

\begin{itemize}
\item {\verb|CML-SER-joint|}: this is the most straightforward strategy that fuses the knowledge learned by CML and SER via minimizing the contrastive loss and self-reconstruction loss jointly, as shown in Fig.~\ref{fusion_strategy}(a). An encoder is shared and jointly optimized by CML and SER, which is used for evaluation after the unsupervised pre-training;
\item {\verb|CML-pretrain-SER|}: we first train CML, then the encoder in SER is initialized with the learned parameters of the query encoder in CML; 
\item {\verb|SER-pretrain-CML|}: we first train SER, then the query encoder and momentum encoder in CML is initialized with the learned parameters of the encoder in SER; 
\item{\verb|SER-distill-CML|}: the learned knowledge of SER is distilled to CML. Namely, this approach is the reverse of CRRL, as shown in Fig.~\ref{fusion_strategy}(b). After the unsupervised pre-training, the query encoder of CML is used for evaluation.
\end{itemize}

The results of different fusion strategies are reported in Table~\ref{fusion-table}. The first three rows are the baselines. Rand-Enc stands for the encoder initialized with random weights. CML denotes that we conduct the unsupervised pre-training with CML alone. Likewise, SER means that we conduct the unsupervised pre-training with SER independently. Table~\ref{fusion-table} shows that CML, SER, and all fusion approaches significantly outperform the Rand-Enc, demonstrating their effectiveness in learning meaningful semantics. 

CML-SER-joint is the first fusion strategy we adopt because joint training is commonly used in multi-task learning. However, as shown in Table~\ref{fusion-table}, CML-SER-joint's performances are only slightly better than SER, indicating that the joint training brings very limited benefit. To explore the reasons, we draw out the training loss, as shown in Fig.~\ref{loss}(a), and find two things: 1) the losses' convergence pace of CML and SER differs significantly. Specifically, the loss of SER reduces rapidly and converges in about 60 epochs, which is much faster than the loss of CML that converges in about 100 epochs; 2) CML-SER-joint achieves worse contrastive loss than CML. For CML-SER-joint, the contrastive loss converges at 2.43, while it is 2.15 when CML is trained independently. This implies that when joint optimizing, SER can affect CML's optimization. 

Next, to avoid the joint optimization problem, we separate the training procedure of CML and SER, proposing CML-pretrain-SER and SER-pretrain-CML. However, as shown in Table~\ref{fusion-table}, the overall performance of CML-pretrain-SER is on bar with SER, indicating that SER does not benefit much from the pre-trained weights of CML. Similarly, SER-pretrain-CML does not compare favorably to CML on most datasets, like NTU RGB+D 120. We speculate that this is because the distribution of the weights differs considerably between the networks trained by CML and SER.

Thirdly, to address these two aforementioned problems, we propose to take advantage of knowledge distillation so that we can implement the information fusion by knowledge transfer, rather than by direct weights inheritance or joint optimization. Specifically, we have two options to implement the knowledge distillation: 1) we first train CML and then freeze it and transfer its knowledge to SER (\emph{i.e.}, CRRL approach); 2) reversely, we first train SER, and then frozen it and transfer its knowledge to CML (\emph{i.e.}, SER-distill-CML approach). As shown in Table~\ref{fusion-table}, CRRL's performances improve considerably over SER, verifying that the motion dynamics learned by CML has been coupled well with the postural information learned by SER. Moreover, the training loss shown in Fig.~\ref{loss}(b) indicates that the reconstruction task and distillation task are learned at a consistent rate. Similarly, SER-distill-CML surpasses CML significantly (\textgreater4.2\% on average), demonstrating the postural information has been distilled to CML well. However, the performance of  SER-distill-CML is still a bit lower than SER. We conjecture that this is because there exists information loss with knowledge distillation since the distillation loss is impossible to reduce to zero during training.


\subsubsection{Velocity-coordinate versus Coordinate-coordinate Contrastive Learning}  \label{vs}

To demonstrate the effectiveness of the proposed velocity-coordinate contrastive learning, we compare it to the traditional coordinate-coordinate contrastive learning. We implement three widely used positional augmentations~\cite{rao2021augmented, yan2018spatial}, which are described as follows.

\begin{itemize}
\item {\verb|Gaussian noise|}: gaussian noise \textit{N}(0, 0.05) is added to the coordinates of skeleton sequences.

\item {\verb|Shear|}: we slant the skeleton sequence to a random angle by using a shear transformation matrix \textit{\textbf{S}}:

\begin{equation}
 \textit{\textbf{S}}=
 \left[
 \begin{matrix}
   1       &   s_{xy}   &  s_{xz} \\
   s_{yx}  &   1        &  s_{yz} \\
   s_{zx}  &   s_{zy}   &  1
  \end{matrix}
  \right],
\end{equation}
where $s_{xy}$, $s_{xz}$, $s_{yx}$, $s_{yz}$, $s_{zx}$, and $s_{zy}$ are shear factors randomly sampled from [-1, 1].

\item {\verb|Rotation|}: following~\cite{rao2021augmented}, we rotate the skeleton sequence by utilizing a general rotation matrix which is obtained from three basic rotation matrices using matrix multiplication. These matrices are defined as the following:

\begin{equation}
 \textit{\textbf{R}}_x(\alpha)=
 \left[
 \begin{matrix}
   1         &   0           &  0 \\
   0         &   cos\alpha   &  sin\alpha \\
   0         &   -sin\alpha  &  cos\alpha
  \end{matrix}
  \right],
\end{equation}

\begin{equation}
 \textit{\textbf{R}}_y(\beta)=
 \left[
 \begin{matrix}
   cos\beta  &   0           &  -sin\beta \\
   0         &   1           &  0 \\
   sin\beta  &   0           &  cos\beta
  \end{matrix}
  \right],
\end{equation}

\begin{equation}
 \textit{\textbf{R}}_z(\gamma)=
 \left[
 \begin{matrix}
   cos\gamma  &   sin\gamma   &  0 \\
   -sin\gamma &   cos\gamma   &  0 \\
   0          &   0           &  1
  \end{matrix}
  \right],
\end{equation}

\begin{equation}
 \textit{\textbf{R}} = \textit{\textbf{R}}_x(\alpha) \textit{\textbf{R}}_y(\beta) \textit{\textbf{R}}_z(\gamma),
\end{equation}

where $\alpha$, $\beta$, $\gamma$ are rotation angles randomly sampled from [$-\pi/4, \pi/4$].

\end{itemize}

The comparison results shown in Table~\ref{contrastive-comparison} demonstrate that the proposed velocity-coordinate contrastive learning outperforms the coordinate-coordinate contrastive learning by large margins on NTU RGB+D 60/120 datasets. We attribute the superior performance of the velocity-coordinate contrastive learning to that compared with positional augmentation (like Gaussian noise, rotation, and shear), velocity sequences are computed as the differences between two consecutive coordinate frames, and thus the absolute spatial coordinates information is removed. Therefore, the model is forced to focus on commonalities in temporal dynamics to maximize the similarity. Namely, velocity directly reflects the motion and inherently contains richer dynamic information, facilitating the velocity-coordinate contrastive learning to extract the motion dynamics.

\begin{figure*}
  \centering
  \subfigure[CML]{
  \includegraphics[width=0.3\linewidth]{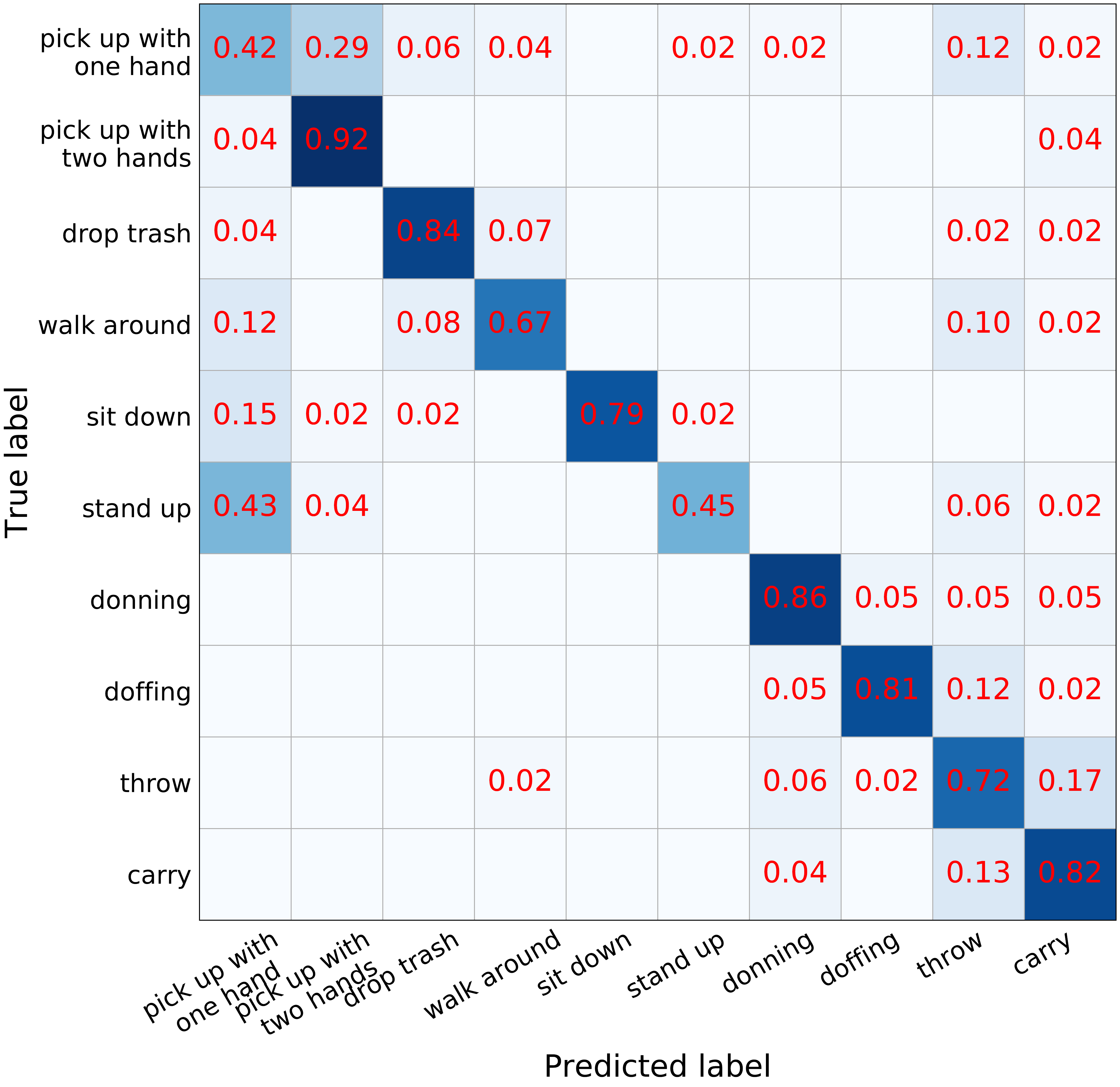}}
  \subfigure[SER]{
  \includegraphics[width=0.3\linewidth]{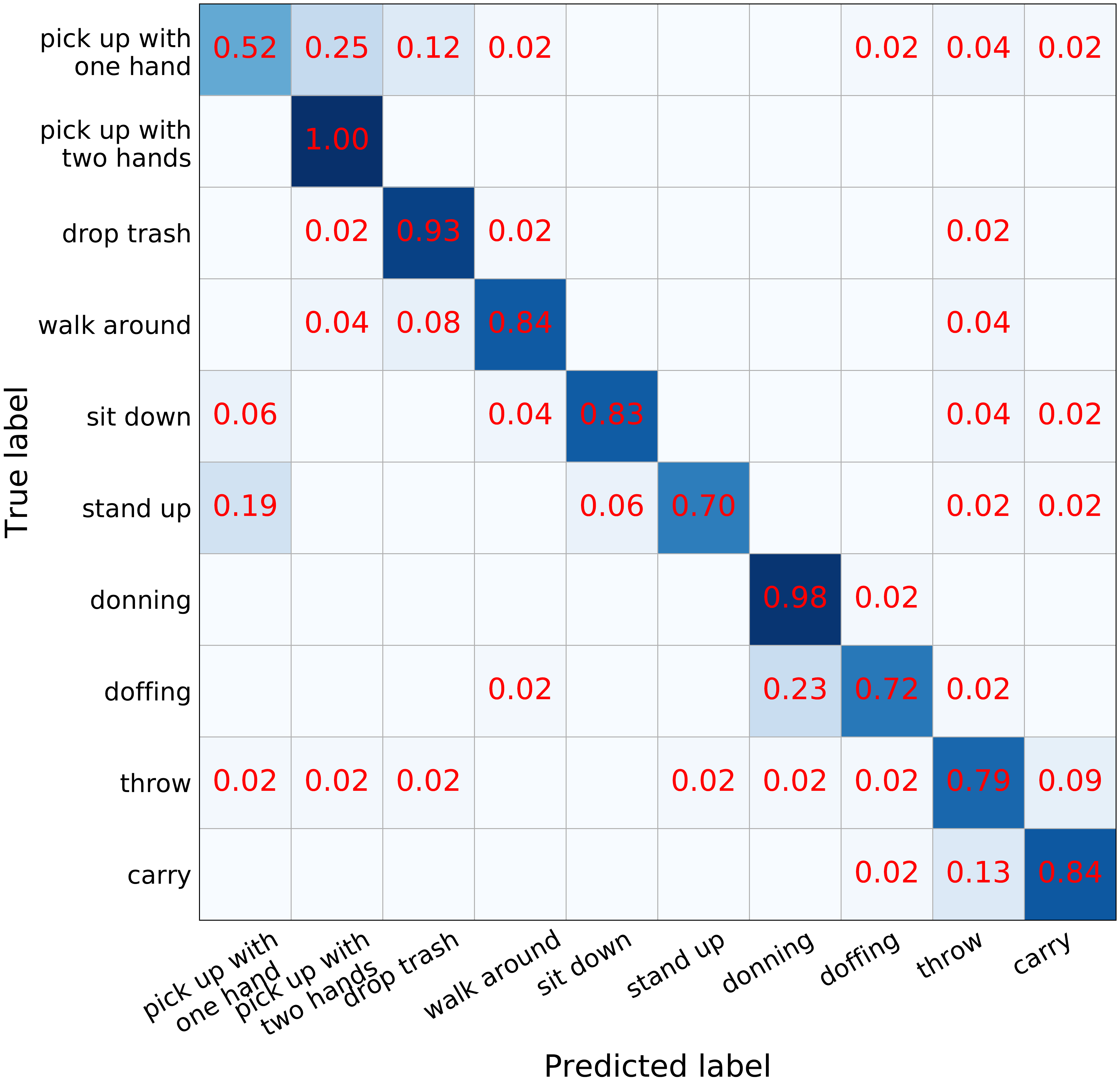}}
  \subfigure[CRRL]{
  \includegraphics[width=0.345\linewidth]{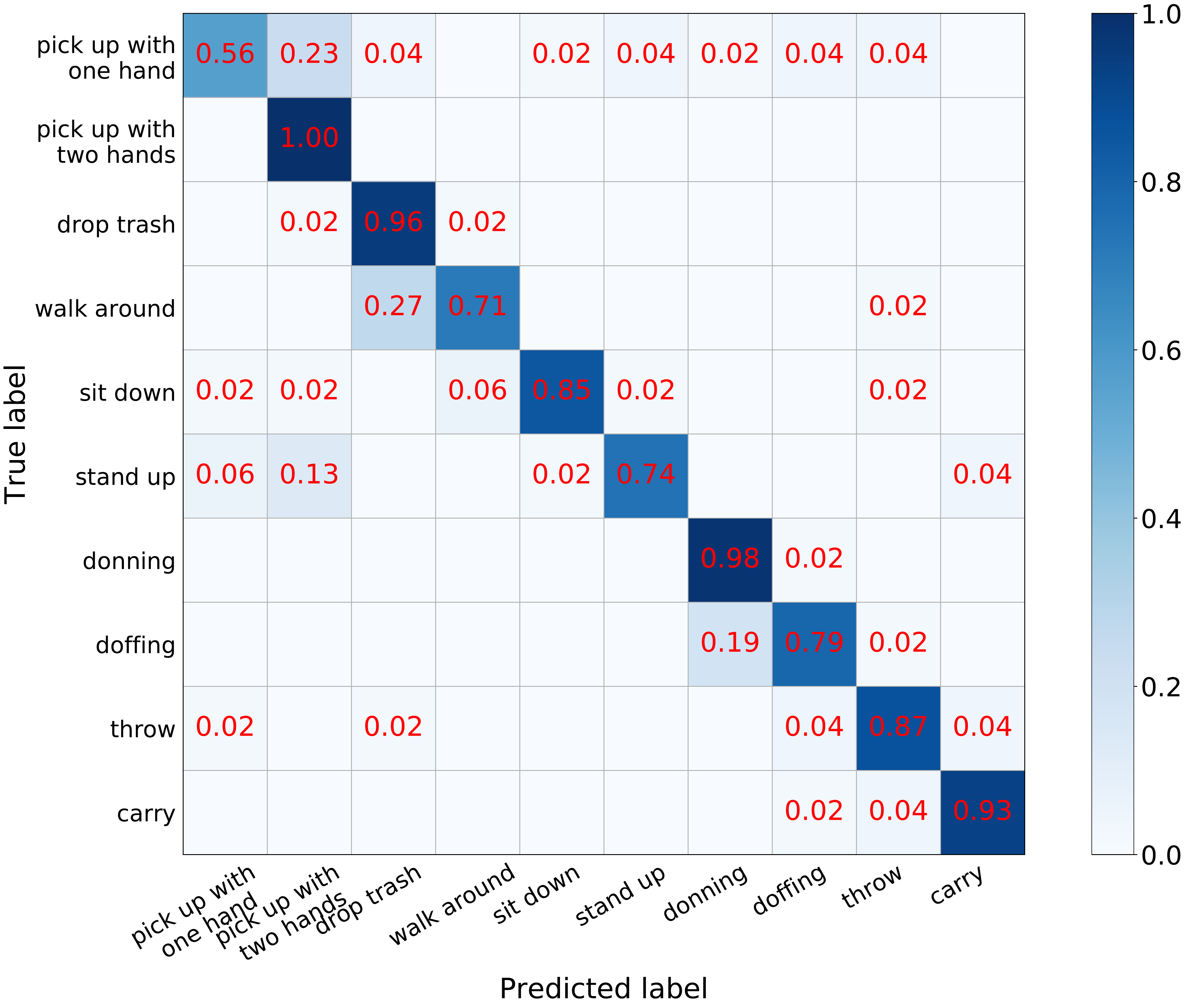}}
  \caption{Confusion matrices of the (a) CML, (b) SER, and (c) CRRL on the NW-UCLA dataset. The numbers less than 0.02 are omitted for clear view.}
  \label{confusion}
\end{figure*}

\begin{figure*}
  \centering
  \subfigure[Coordinates distributions]{
  \includegraphics[width=0.4\linewidth]{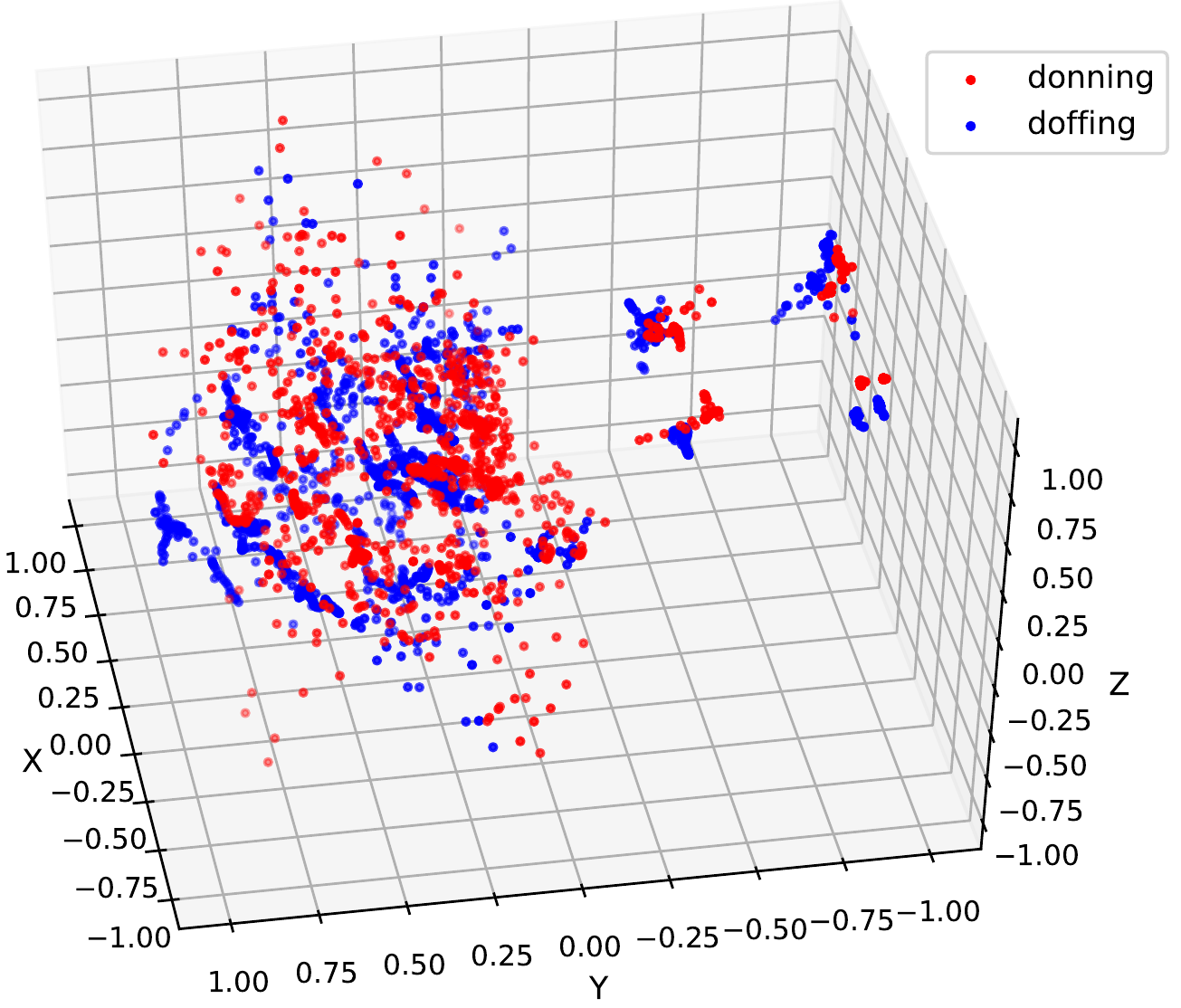}}
  \subfigure[Velocity distributions]{
  \includegraphics[width=0.4\linewidth]{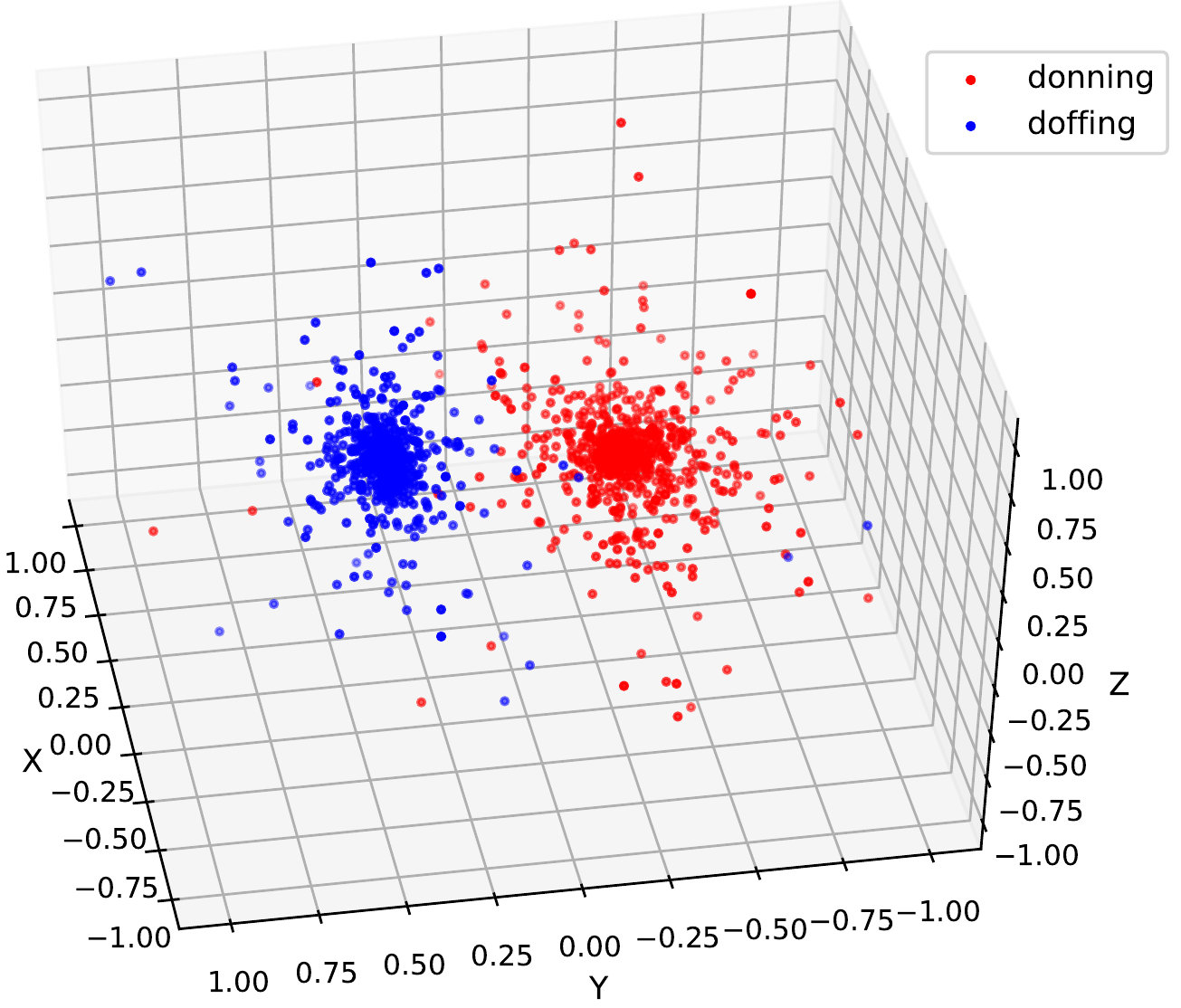}}
  \caption{The coordinates and velocity distributions of ``donning'' (red) and ``doffing'' (blue). While ``donning'' and ``doffing'' have similar spatial coordinates distributions, their velocity distributions are dissimilar. Technically, we randomly select a sample from the ``donning'' samples of the test set of the NW-UCLA dataset, and likewise a sample is randomly selected from the ``doffing'' samples. Then we present the selected samples' skeleton coordinates sequence and their velocity sequence in (a) and (b), respectively.}
  \label{distribution}
\end{figure*}

\begin{figure*}
  \centering
  \subfigure[CML]{
  \includegraphics[width=0.282\linewidth]{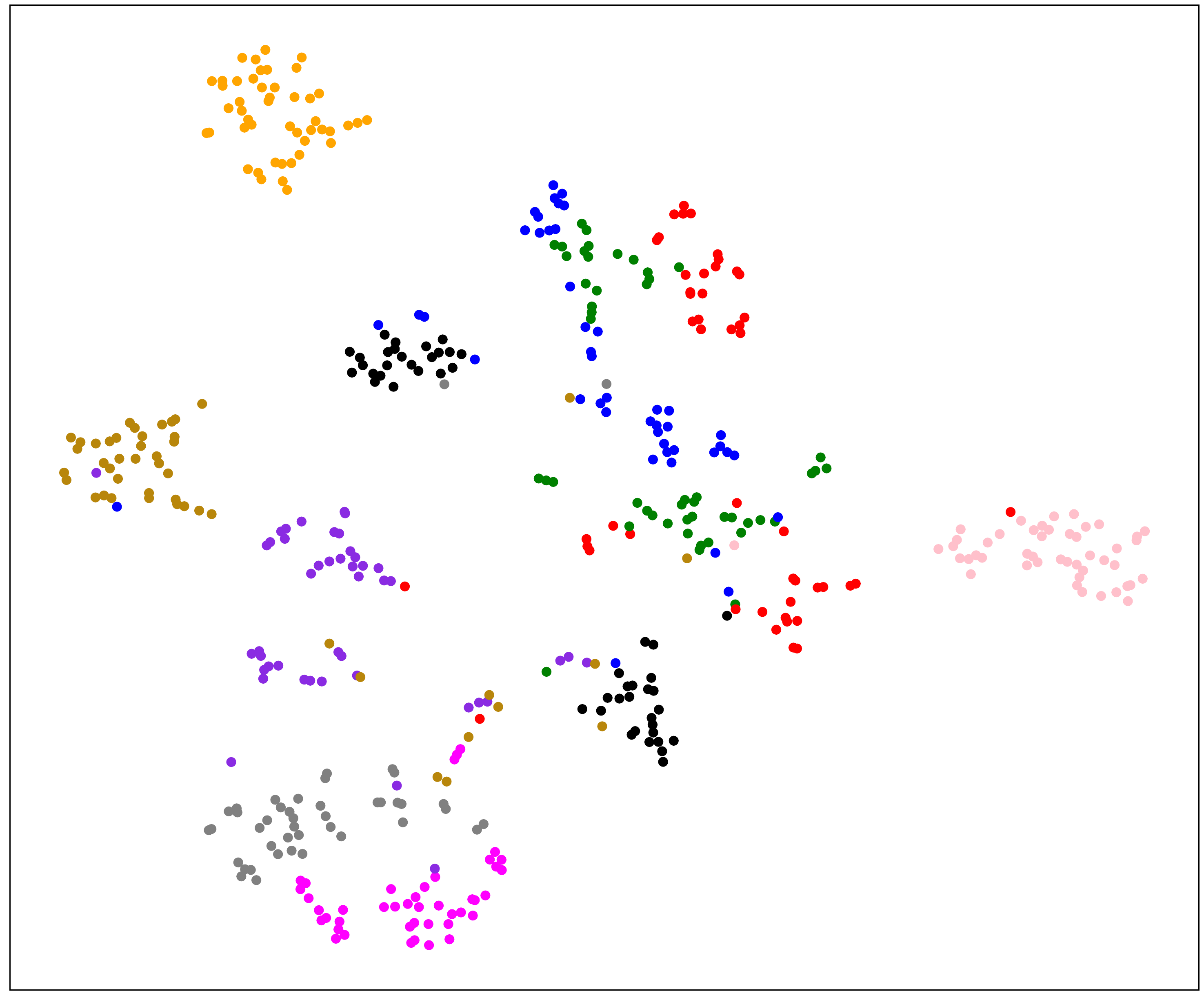}}
  \subfigure[SER]{
  \includegraphics[width=0.282\linewidth]{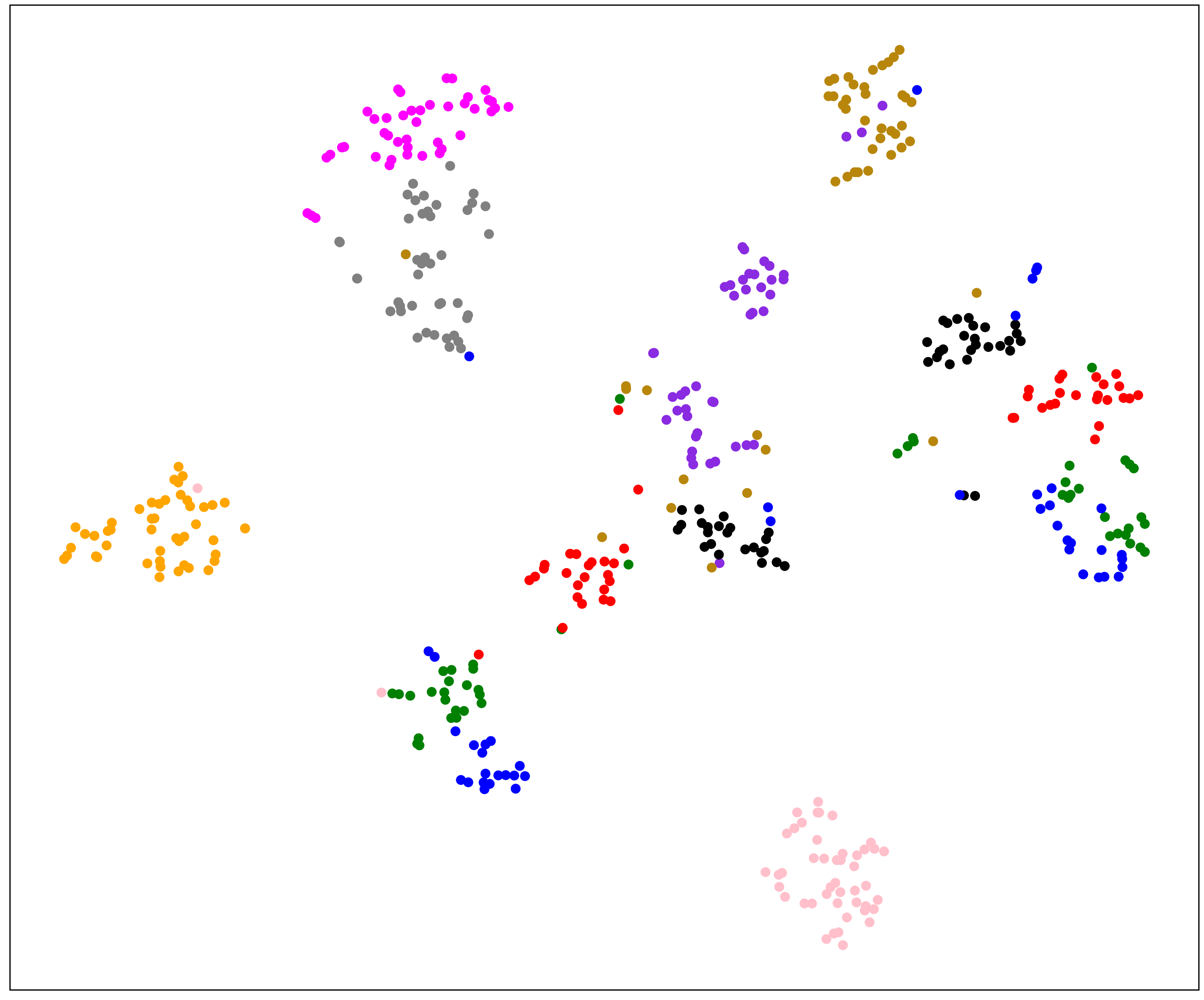}}
  \subfigure[CRRL]{
  \includegraphics[width=0.282\linewidth]{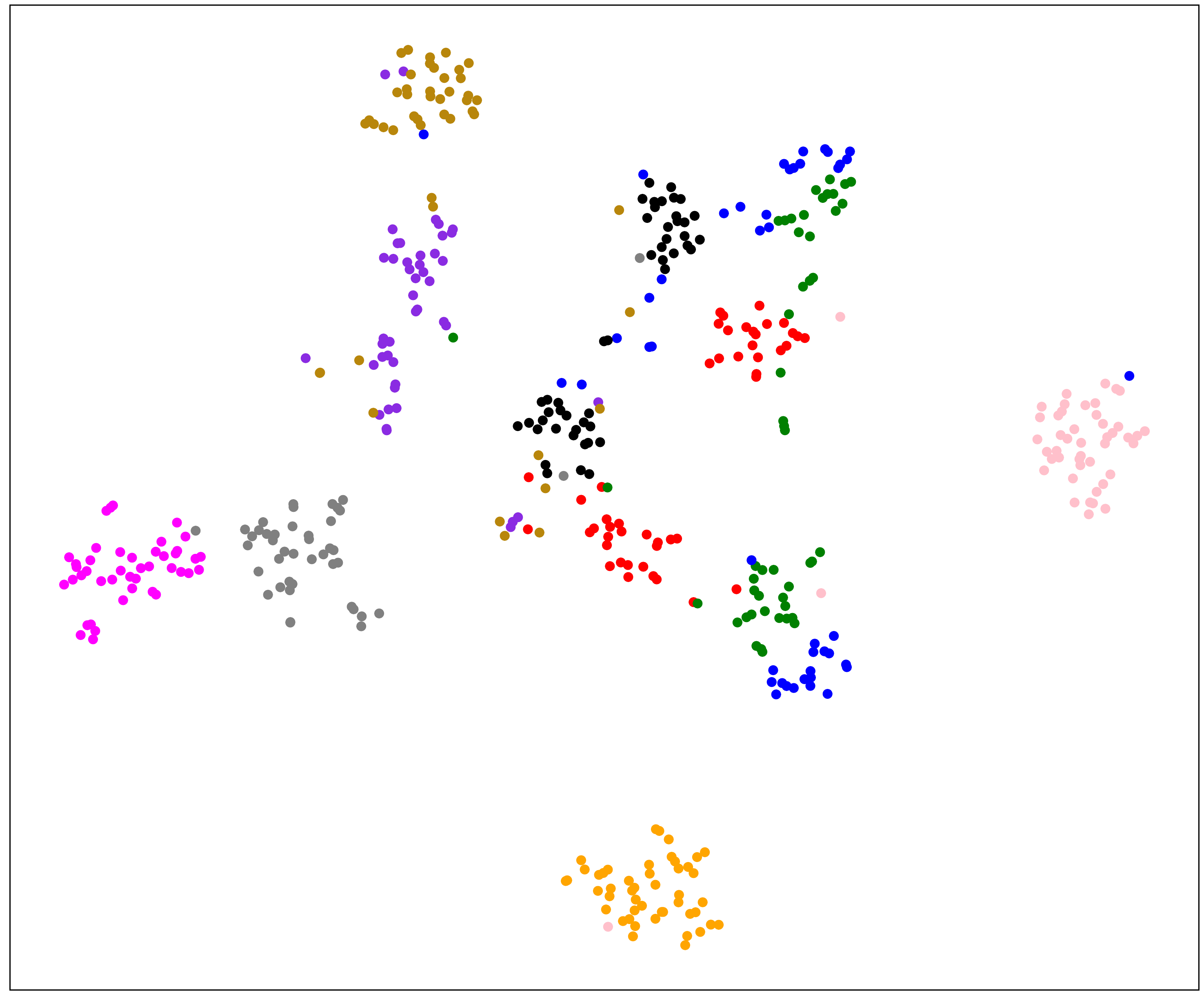}}
  \subfigure{
  \includegraphics[width=0.0968\linewidth]{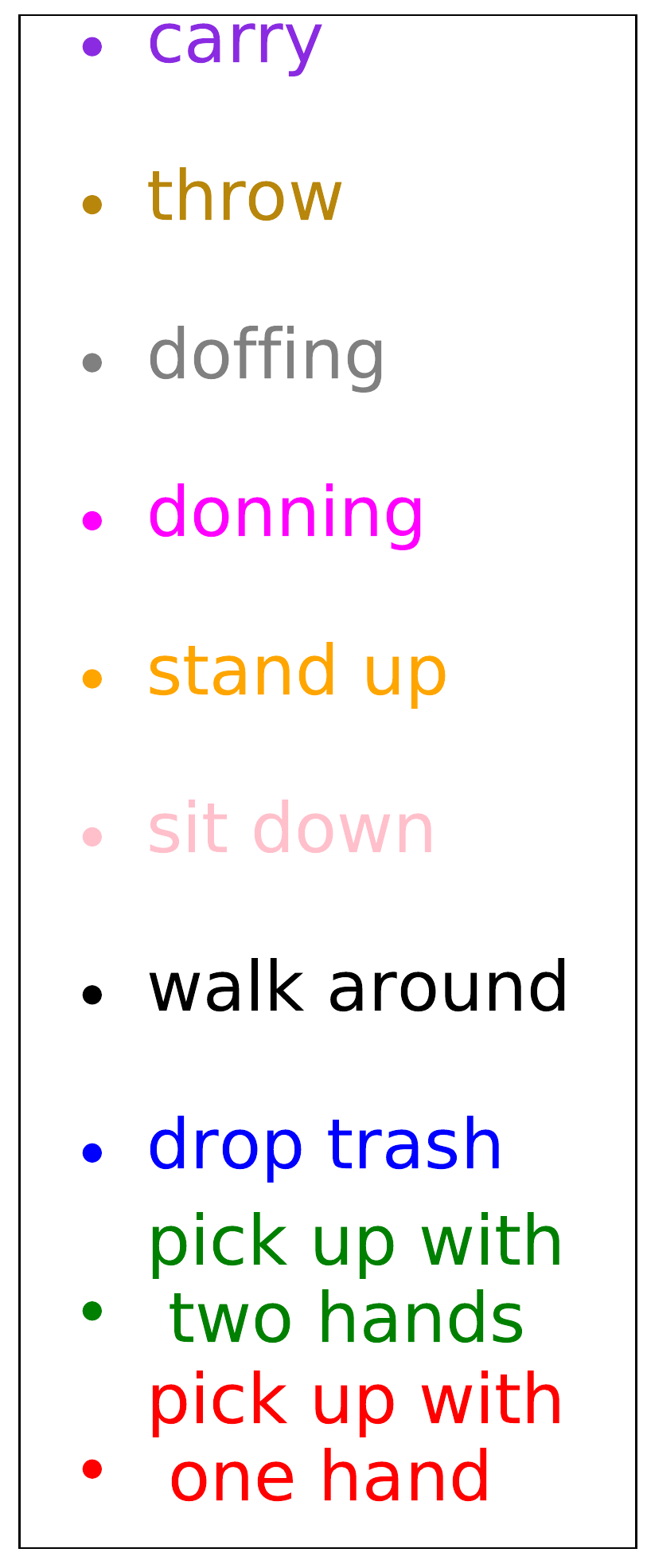}}
  \caption{t-SNE visualizations of the hidden vectors of (a) CML, (b) SER, and (c) CRRL on the test set of NW-UCLA dataset. The bar on the right describes the one-to-one correspondence between color and action class.}
  \label{tsne}
\end{figure*}


\subsubsection{Visualizations}   \label{visual}

To verify the effectiveness of the combination of CML and SER, we visualize the action classification results of CML, SER, and CRRL so that we can conduct a more detailed analysis of their learned representations. Specifically, we present their confusion matrices and t-SNE~\cite{van2008visualizing} visualizations in this section. We implement these experiments on the NW-UCLA dataset since this dataset's number of action categories is suitable for visualization.  

Fig.~\ref{confusion} shows the confusion matrices of CML, SER, and CRRL. For almost every action, CRRL's classification accuracy is higher than either CML or SER, verifying that the combination of CML and SER improves the representation of every action. For example, the classification accuracy of ``carry'' is improved from 82\% and 84\% to 93\%.

Furthermore, ``donning'' and ``doffing'' is very similar in postural view (see Fig.~\ref{distribution}(a)), so that SER misclassifies 23\% ``doffing'' as ``donning''. However, these two action is different in velocity view (see Fig.~\ref{distribution}(b)), since they are temporally inverse to each other, so that CML misclassifies only 5\% ``doffing'' as ``donning''. This result demonstrates that SER tends to focus on postural information, while CML can capture temporal dynamics in representation learning. This conclusion is further verified by ``sit down'',  ``stand up'', and other actions.

\begin{table}
  \caption{Performance comparison of different reconstruction methods.}
  \label{recon-both-direction-table}
  \setlength{\tabcolsep}{2.95mm}{
  \centering
  \begin{tabular}{l|cc|cc}
    \hline
    \multirow{2}{*}{Methods}
    & \multicolumn{2}{c|}{NTU 60}
    & \multicolumn{2}{c}{NTU 120}\\
    \cline{2-5}
    & C-View    & C-Sub    & C-Setup   & C-Sub   \\
    \hline
    forward   & 70.7   & 63.9   & 53.6   & 52.6     \\
    reverse   & 71.1   & 64.4   & 51.7   & 51.4     \\
    SER(forward+reverse)  & \textbf{72.2}  & \textbf{65.8}  & \textbf{55.2} & \textbf{54.1} \\
  \hline
\end{tabular}}
\end{table}

Fig.~\ref{tsne} shows t-SNE visualizations of the hidden vectors of CML, SER, and CRRL. It clearly shows that the representations for each class are well separated. In particular, CRRL's representations are better separated than CML and SER.


\subsubsection{Computational Cost} \label{computational_costs}

We estimate the computational cost of the proposed CML, SER, and CRRL. The computational cost is measured in terms of the number of network parameters and the number of floating point operations (FLOPs) computed on a 60-frame action sequence performed by two persons, with 25 joints for each person. The results are shown in Table~\ref{compute-cost-table}. Since CRRL contains two training steps, with step 1 for training CML and step 2 for training SER and INF, its computational cost is higher than either CML or SER.

\begin{table}[t]
  \caption{Computational cost of CML, SER, and CRRL.}
  \label{compute-cost-table}
  \centering
  \begin{tabular}{p{2cm}p{1.5cm}p{1.5cm}p{1.5cm}}
    \hline
                     & CML     & SER     & CRRL     \\
    \hline
    parameters (M)   & 4.7     & 2.2     & 7.4      \\
    FLOPs (M)        & 438.1   & 270.4   & 927.5    \\
  \hline
\end{tabular}
\end{table}


\subsubsection{Ablation Study of SER Module} \label{Analysis_SER_section}

To verify the effectiveness of the proposed reconstruction strategy, \emph{i.e.}, reconstructing the input skeleton sequence both forwardly and reversely, we compare it to the following two approaches:
\begin{itemize}
  \item [$\bullet$]the auto-encoder that only reconstructs the input skeleton sequence forwardly. We denote this approach as \emph{forward}.
  \item [$\bullet$] the auto-encoder that only reconstructs the input skeleton sequence reversely. We denote this approach as \emph{reverse}.
\end{itemize}

Table~\ref{recon-both-direction-table} shows the comparison between them. The performance of \emph{forward} and \emph{reverse} is almost on par with each other, and SER surpasses either of them by nontrivial margins. For instance, on NTU RGB+D 120 dataset under cross-subject protocol, SER's accuracy is 54.1\%, higher than either \emph{forward} (52.6\%) or \emph{reverse} (51.4\%). These results support our motivation to reconstruct the input sequence both forwardly and reversely.


\begin{table*}
  \caption{The influence of the momentum coefficient on CML module w.r.t. action classification accuracy on NTU RGB+D 60 dataset with cross-subject protocol. $m_{enc}$ denotes the momentum for the key encoder, and $m_{mlp}$ denotes the momentum for the key MLP.}
  \label{momentum-table}
  \centering
  \setlength{\tabcolsep}{4.1mm}{
  \begin{tabular}{c|c|c|l}
    \hline
    $m_{enc}$   & $m_{mlp}$   & Accuracy(\%)  & Remarks \\
    \hline
    1.0             & 1.0             & 55.9          & Both the key encoder and the key MLP are not updated. \\
    \hline
    \multirow{4}{*}{0.999}
    & 0.999      & 50.5  & \multirow{3}{*}{The key encoder and the key MLP are updated with}\\
    & 0.9999     & 51.9  & \multirow{3}{*}{independent momentum.}\\
    & 0.99999    & 58.1  \\
    & 1.0        & \textbf{62.8}  \\
    \hline
    0.5     & \multirow{7}{*}{1.0} & 60.1    & \multirow{7}{*}{The key MLP is not updated.} \\
    0.7     & \multirow{7}{*}{}    & 61.2 \\
    0.8     & \multirow{7}{*}{}    & 62.0 \\
    0.9     & \multirow{7}{*}{}    & 62.3 \\
    0.99    & \multirow{7}{*}{}    & 62.5 \\
    0.999   & \multirow{7}{*}{}    & \textbf{62.8} \\
    0.9999  & \multirow{7}{*}{}    & 59.5 \\

    \hline
\end{tabular}}
\end{table*}

\begin{figure*}
  \centering
  \subfigure[The influence of the number of negative samples $K$]{
  \includegraphics[width=0.48\linewidth]{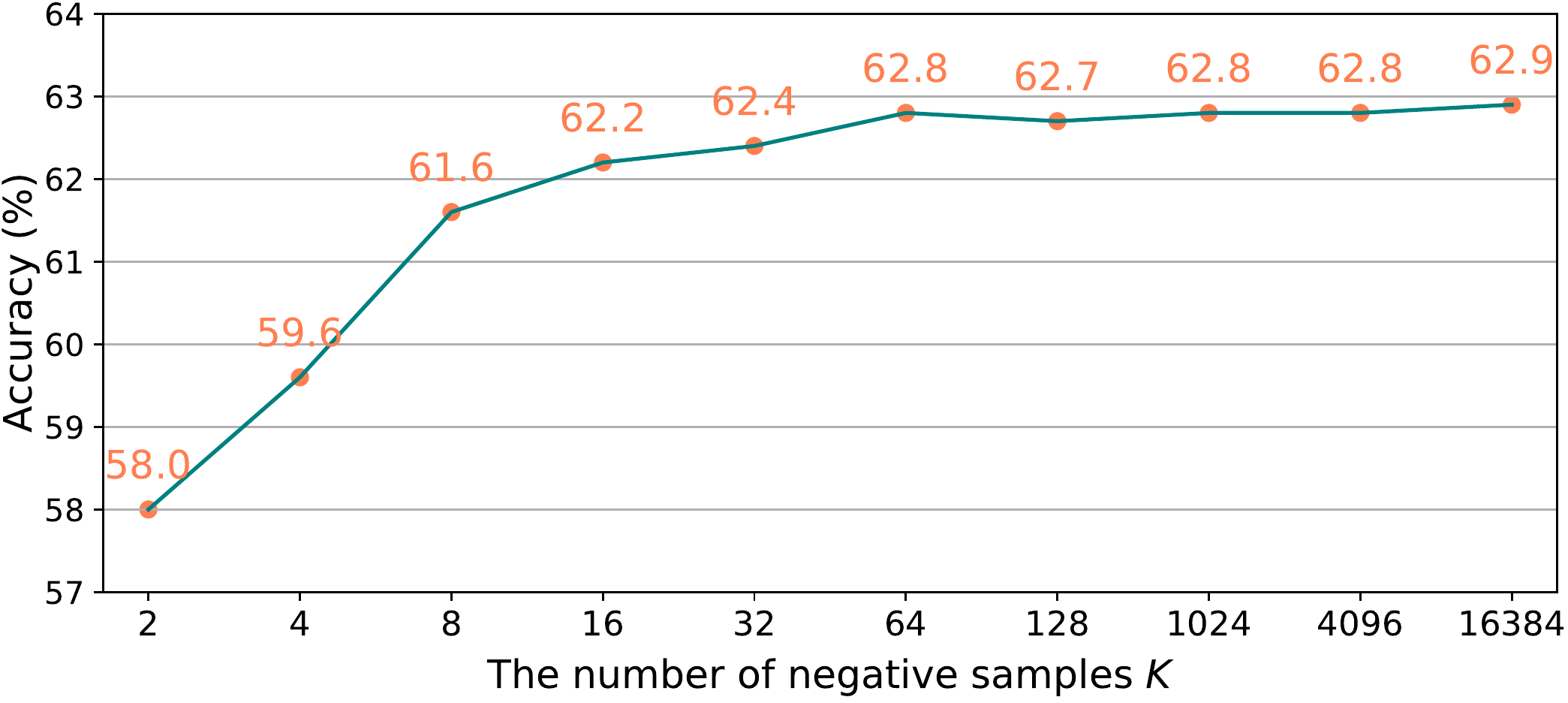}}
  \subfigure[The influence of temperature $\tau$]{
  \includegraphics[width=0.48\linewidth]{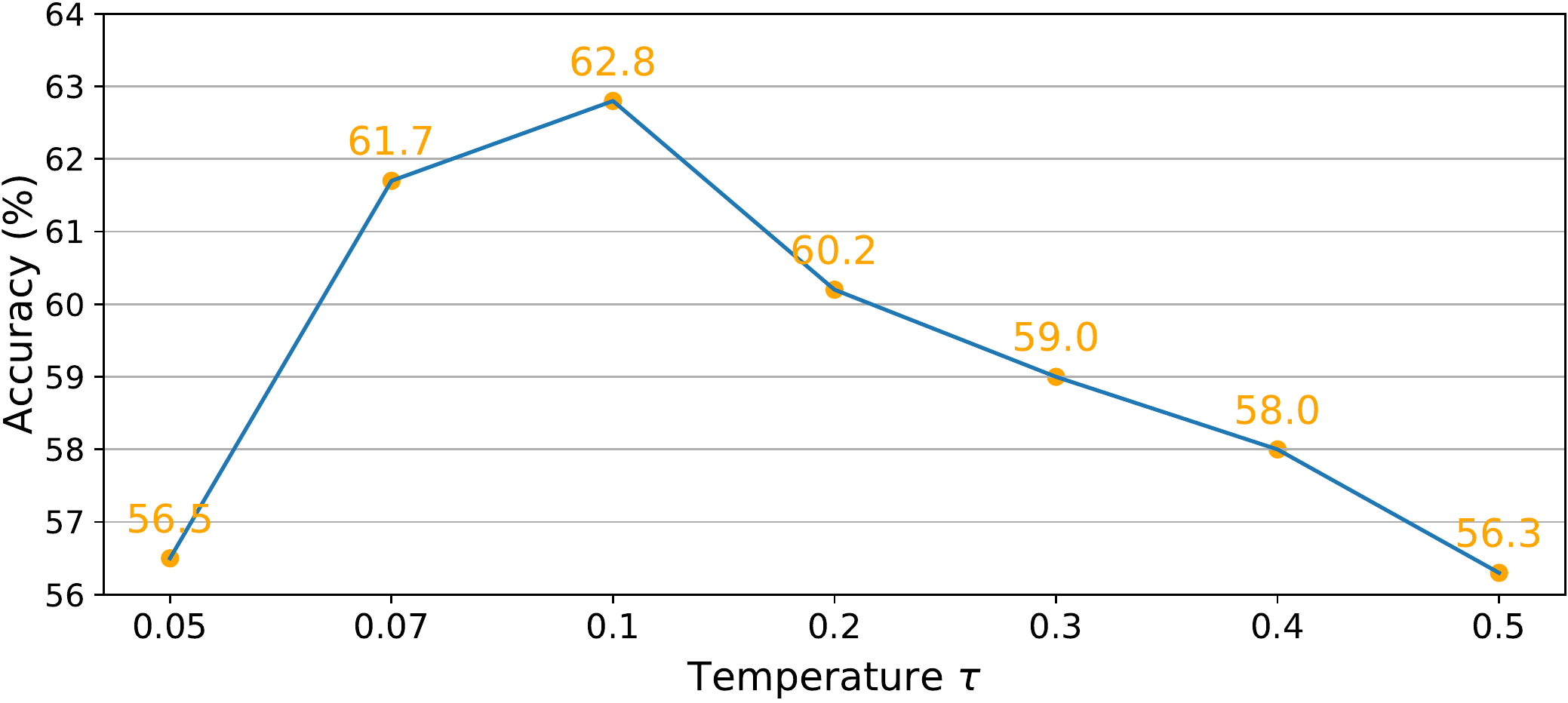}}
  \caption{The influence of the number of negative samples $K$ and temperature $\tau$ on CML module w.r.t. action classification accuracy on NTU RGB+D 60 dataset with cross-subject protocol.}
  \label{parameter}
\end{figure*}

\subsubsection{Ablation Study of CML Module} \label{Analysis_CML_section}

In this section, we report the ablation studies of the CML module with respect to hyper-parameters, \emph{i.e.}, the number of negative samples $K$, momentum $m$, and temperature $\tau$. We conduct the experiments on NTU RGB+D 60 dataset with the cross-subject protocol.

\textbf{Momentum}. Momentum controls the update rate of the key encoder and the key MLP (\emph{i.e.}, the MLP on top of the key encoder). We notice that in both~\cite{chen2020mocov2} and~\cite{chen2020simple}, the key encoder and the key MLP are updated with the same momentum, but we empirically find that this setting is not optimal for skeleton-based action representation learning. In this section, we study that the key encoder and the key MLP are updated with different momentum. For convenience, we denote the momentum for key encoder as $m_{enc}$, and the momentum for key MLP as $m_{mlp}$. The experimental results are shown in Table~\ref{momentum-table}.

Firstly, as shown in the first row of Table~\ref{momentum-table}, if both the key encoder and key MLP are never updated and remain its random initialization, \emph{i.e.}, $m_{enc}=m_{mlp}=1.0$, the classification accuracy is 55.9\%, which is higher than the Rand-Enc (45.5\%). This result is in line with~\cite{grill2020bootstrap} which found that even using a fixed randomly initialized network as the key encoder can help the query encoder learn meaningful representation.

Next, we make the key encoder updated with $m_{enc}=0.999$ and try to update the key MLP with momentum from 0.999 to 0.99999, see the second group (rows) of Table~\ref{momentum-table}. We observe that slower progressing key MLP is beneficial for higher performance. if $m_{mlp}=1.0$, \emph{i.e.}, the key MLP is never updated, CML module achieves best performance. We speculate that this is because, compared with the progressing key MLP, the not updated random key MLP maps the feature vector of the key encoder to a more random position in the embedding space, reducing the mutual information between contrastive views and improving downstream classification accuracy~\cite{tian2020makes}. To some extent, the not updated key MLP can be considered as a kind of data augmentation.

Then, we keep the key MLP not updated and try to update the key encoder with different momentums, from 0.5 to 0.9999. As shown in the third group of Table~\ref{momentum-table}, updating the key encoder too often and too slowly can harm CML's performance. When the $m_{enc}$ is around $0.99\sim0.999$, it yields the best results.

\textbf{The number of negative samples $K$}. In contrastive methods, the negative samples prevent the model from finding collapsed solutions~\cite{grill2020bootstrap}, \emph{i.e.}, outputting the same feature vectors for all input instances. We show the impact of $K$ in Fig.~\ref{parameter}(a). The model's performance gets improved when $K$ increases from 2 to 64, then the benefit brought by more negative samples is saturated at $K=64$ (for convenience, we denote this number as saturated number). We notice that in~\cite{He2020MomentumCF}, the benefit brought by more negative samples is saturated at around $K=16384$. We conjecture that there are two reasons why our saturated number is much smaller than~\cite{He2020MomentumCF}'s. Firstly, in~\cite{He2020MomentumCF}, the data samples are images with the size of $224\times224\times3$, while in our framework, the data samples are skeleton sequences with much smaller size, \emph{i.e.}, $60\times50\times3$. The smaller size of input data means the smaller variety between different samples, resulting in that it is more likely to find negative examples in small $K$ that are sufficiently close to positive examples to make the contrastive task challenging. This leads to high performance even $K$ is small. Secondly, in~\cite{He2020MomentumCF}, the model size is about 25.5M (ResNet-50), while our model size is an order of magnitude smaller than it, \emph{i.e.} about 2.3M. We speculate that the larger model needs more negative samples for the enhanced representation.

\textbf{Temperature $\tau$}. This parameter controls the concentration level of the similarity distribution~\cite{hinton2015distilling, Wu2018UnsupervisedFL}. Using a higher value for $\tau$ produces a softer similarity distribution over the positive pair and negative pairs. Fig.~\ref{parameter}(b) shows the impact of $\tau$. When $\tau$ increases, the model's performance gets improved first and then drops, and the best performance is achieved when $\tau=0.1$.

\begin{figure*}
  \centering
  \subfigure[Velocity sequence generation]{
  \includegraphics[width=0.48\linewidth]{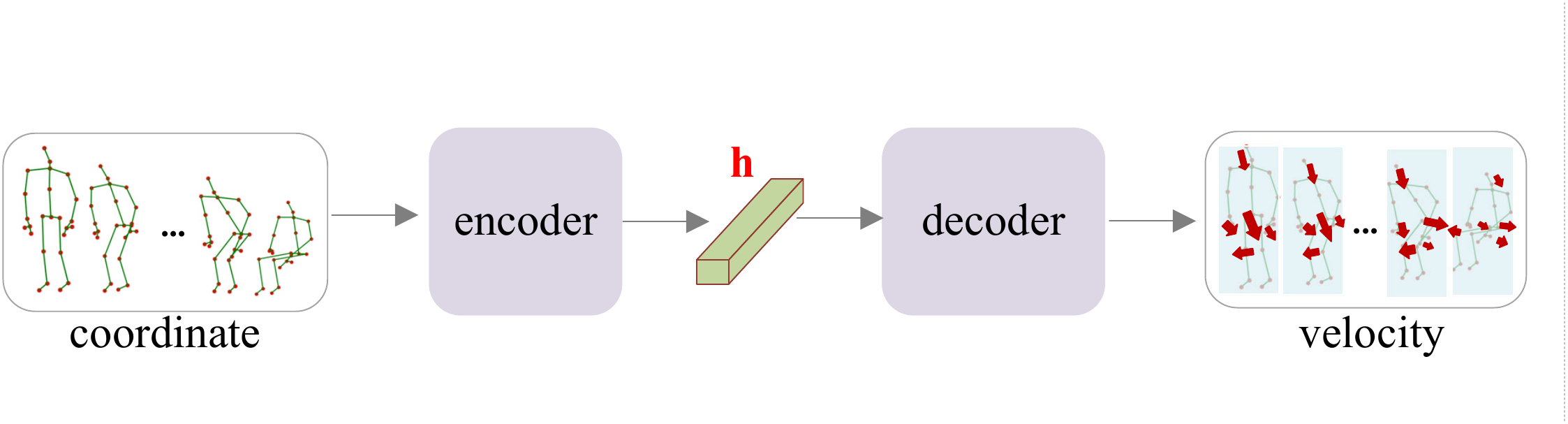}}
  \subfigure[Velocity sequence generation and skeleton sequence reconstruction]{
  \includegraphics[width=0.48\linewidth]{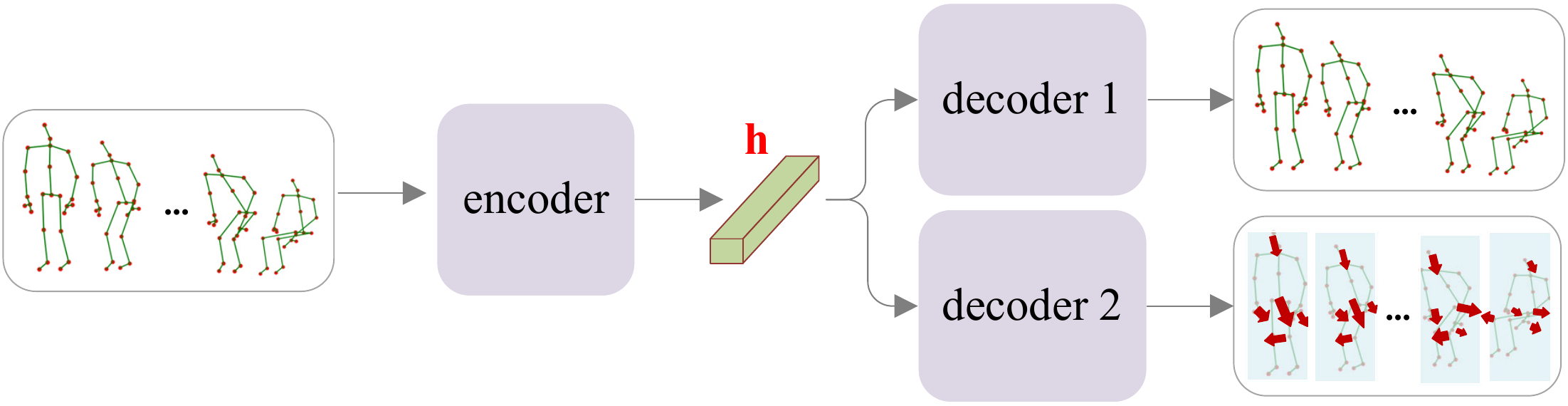}}
  \caption{The illustration of (a) Velocity sequence Generation (VG) framework, and (b) Velocity sequence Generation and Skeleton sequence Reconstruction (VGSR) framework.}
  \label{velocity_recon}
\end{figure*}


\subsubsection{Reconstructing Velocity Sequence}   \label{VG} 

In this section, we propose another approach to guide the model to learn motion dynamics: given the skeleton sequence, the model learns to generate the velocity sequence. We compare this approach with CML and CRRL. The proposed framework is shown in Fig.~\ref{velocity_recon}(a). The RNN based encoder runs through an input skeleton sequence and encodes it to a representation vector, which is then utilized by the decoder to produce the velocity sequence. Since the decoder's generation target is the velocity sequence, the representation learned by the encoder is thus induced to focus on the motions of the input. We denote this method as Velocity sequence Generation (VG). We notice that the very recently published paper~\cite{cheng2021hierarchical} proposes a similar idea that the model learns to predict the motion direction of masked frames, but they present their model based on the hierarchical Transformer.

\begin{table}[t]
  \caption{The performance of VG and VGSR, and comparison to CML and CRRL.}
  \label{velocity-table}
  \centering
  \setlength{\tabcolsep}{3.9mm}{
  \begin{tabular}{l|cc|cc}
    \hline
    \multirow{2}{*}{Methods}
    & \multicolumn{2}{c|}{NTU 60}
    & \multicolumn{2}{c}{NTU 120}\\
    \cline{2-5}
    & C-View    & C-Sub    & C-Setup   & C-Sub   \\
    \hline
    VG                & 65.9   & 61.1   & 52.1   & 50.8     \\
    CML                & 68.1   & 62.8   & 51.6   & 50.9     \\
    \hline
    VGSR               & 72.4   & 66.2   & 55.7   & 55.1    \\
    CRRL  & \textbf{73.8}  & \textbf{67.6}  & \textbf{57.0} & \textbf{56.2} \\
  \hline
\end{tabular}}
\end{table}

Further, we combine the velocity sequence generation task with the skeleton sequence reconstruction task via joint optimization, as shown in Fig.~\ref{velocity_recon}(b), so that the model can simultaneously learn the postures and motions. We denote this model as Velocity sequence Generation and Skeleton sequence Reconstruction (VGSR).

We present the performances of VG and VGSR on NTU 60/120 RGB-D datasets in Table~\ref{velocity-table} and compare them with CML and CRRL. It shows that the overall performance of VG does not compare favorably to CML, indicating that the VG approach might not be as effective as CML in learning motion dynamics. The final rows of Table~\ref{velocity-table} present the inferior performance of VGSR compared to CRRL, showing that the latter is stronger in capturing postures and motions.


\section{Conclusions}

In this paper, we proposed a Contrast-Reconstruction Representation Learning framework for self-supervised skeleton-based action recognition, which contains three modules: 1) \emph{Sequence Reconstructor}; 2) \emph{Contrastive Motion Learner}; and 3) \emph{Information Fuser}. The \emph{Sequence Reconstructor} learns postural information via coordinate sequence reconstruction. The \emph{Contrastive Motion Learner} captures the motion dynamics by maximizing the similarity between the representations learned from the velocity sequence and the skeleton sequence. Finally, the \emph{Information Fuser} couples the learned posture information and motion dynamics via knowledge distillation. Experiments on five publicly available datasets demonstrated the substantial improvements of our proposed method over state-of-the-art models using RNN architectures. Moreover, in experiments, we showed that: 1) compared with only re-generating the input sequence in one direction, the reconstruction-based learning benefits from re-producing the input sequence both forwardly and reversely; 2) compared with momentum updated key MLP, the contrastive learning benefits from not updated key MLP. In the future, we will explore more possibilities of combining reconstruction-based learning and contrastive learning for unsupervised human action recognition.


\ifCLASSOPTIONcaptionsoff
  \newpage
\fi

\bibliographystyle{IEEEtran}
\bibliography{IEEEabrv}

\end{document}